\documentclass{article} %
\usepackage{iclr2024_conference,times}

\usepackage{amsmath,amsfonts,bm}

\def\Figref#1{Figure~\ref{#1}}

\def\Secref#1{Section~\ref{#1}}

\def\eqref#1{equation~\ref{#1}}
\def\Eqref#1{Equation~\ref{#1}}

\def\1{\bm{1}}

\DeclareMathAlphabet{\mathsfit}{\encodingdefault}{\sfdefault}{m}{sl}
\SetMathAlphabet{\mathsfit}{bold}{\encodingdefault}{\sfdefault}{bx}{n}

\usepackage[hidelinks,colorlinks]{hyperref}
\usepackage{url}

\usepackage{graphicx}
\usepackage{amssymb}
\newcommand{\mb}[1]{\mathbf{#1}}

\usepackage{enumitem}
\usepackage[normalem]{ulem}

\newcommand{\uit}[1]{\uline{\textit{#1}}}

\usepackage{multirow}
\usepackage{booktabs}
\usepackage{color}
\usepackage{xspace}
\usepackage{wrapfig}
\usepackage{subcaption}

\definecolor{red}{rgb}{0.8,0,0}  
\definecolor{green}{RGB}{0, 133, 21}  
\definecolor{grey}{rgb}{0.5,0.5,0.5}  

\makeatletter
\def\blfootnote{\xdef\@thefnmark{}\@footnotetext}
\DeclareRobustCommand\onedot{\futurelet\@let@token\@onedot}
\def\@onedot{\ifx\@let@token.\else.\null\fi\xspace}
\def\eg{\textit{e.g}\onedot}

\def\ie{\textit{i.e}\onedot}

\makeatother

\usepackage[
  tableposition = bottom
]{caption}
\usepackage[title]%
  {appendix}
\def\Tabref#1{Table~\ref{#1}}

\newcommand{\methodname}{\textsc{MagicDrive}\xspace}

\title{\methodname: Street View Generation with \\Diverse 3D Geometry Control}

\author{Ruiyuan Gao$^{1*}$,
Kai Chen$^{2*}$,
Enze Xie$^{3\dag}$,
Lanqing Hong$^{3}$, \\
\textbf{Zhenguo Li}$^{3}$
\textbf{, Dit-Yan Yeung}$^{2}$
\textbf{, Qiang Xu}$^{1\dag}$\\
$^{1}$The Chinese University of Hong Kong
\enspace
$^{2}$Hong Kong University of Science and Technology \\
$^{3}$Huawei Noah's Ark Lab\\
\texttt{\{rygao,qxu\}@cse.cuhk.edu.hk},\enspace\texttt{kai.chen@connect.ust.hk},\enspace\\
\texttt{\{xie.enze,honglanqing,li.zhenguo\}@huawei.com},\enspace\texttt{dyyeung@cse.ust.hk} \\
}

\iclrfinalcopy %
\begin{document}

\maketitle
\vspace{-0.5cm}
\begin{abstract}

\vspace{-0.2cm}
Recent advancements in diffusion models have significantly enhanced the data synthesis with 2D control.
Yet, precise 3D control in street view generation, crucial for 3D perception tasks, remains elusive.
Specifically, utilizing Bird's-Eye View (BEV) as the primary condition often leads to challenges in geometry control (\eg, height), affecting the representation of object shapes, occlusion patterns, and road surface elevations, all of which are essential to perception data synthesis, especially for 3D object detection tasks.
In this paper, we introduce \methodname, a novel street view generation framework, offering diverse 3D geometry controls including camera poses, road maps, and 3D bounding boxes, together with textual descriptions, achieved through tailored encoding strategies. 
Besides, our design incorporates a cross-view attention module, ensuring consistency across multiple camera views. With \methodname, we achieve high-fidelity street-view image \& video synthesis that captures nuanced 3D geometry and various scene descriptions, enhancing tasks like BEV segmentation and 3D object detection.
\blfootnote{
$^{*}$Equal contribution.
$^{\dag}$Corresponding authors.
Project Page: \scriptsize{\url{https://flymin.github.io/magicdrive}}.
}

\end{abstract}

\begin{figure}[ht]
    \centering
    \vspace{-0.6cm}
    \includegraphics[width=0.98\linewidth]{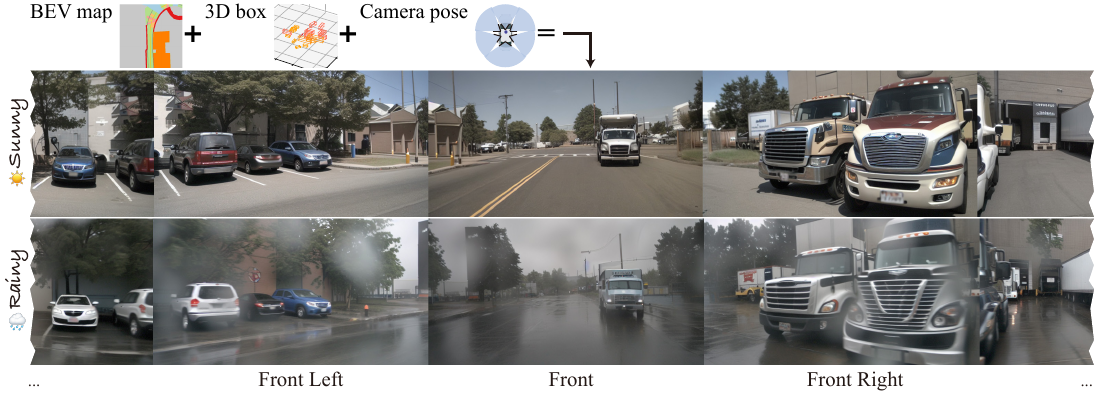}
    \vspace{-0.4cm}
    \caption{
    Multi-camera street view generation from \methodname. 
    \methodname can generate continuous camera views with controls from the road map, object boxes, and text (\eg, weather).
    }
    \label{fig:teaser}
    \vspace{-0.7cm} 
\end{figure}

\section{Introduction}
\vspace{-0.1cm}
The high costs associated with data collection and annotation often impede the effective training of deep learning models. 
Fortunately, cutting-edge generative models have illustrated that synthetic data can notably boost performance across various tasks, such as object detection~\citep{chen2023integrating} and semantic segmentation~\citep{wu2023datasetdm}. 
Yet, the prevailing methodologies are largely tailored to 2D contexts, primarily relying on 2D bounding boxes~\citep{lin2014microsoft,han2021soda10m} or segmentation maps~\citep{ADE20K} as layout conditions~\citep{chen2023integrating,li2023gligen}.

\begin{figure}[t]
    \centering
    \includegraphics[width=0.98\linewidth]{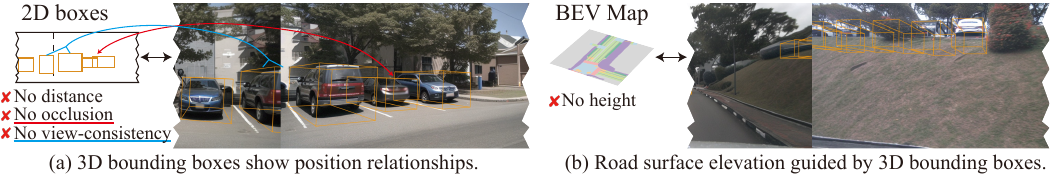}
    \vspace{-0.3cm}
    \caption{
    3D bounding boxes are crucial for street view synthesis.
    Two examples show that 2D boxes or BEV maps lost distance, height, and elevation. Images are generated from \methodname.}
    \label{fig:teaser2}
    \vspace{-0.5cm}
\end{figure}

In autonomous driving applications, a thorough grasp of the 3D environment is essential. This demands reliable techniques for tasks like Bird’s-Eye View (BEV) map segmentation~\citep{zhou2022cross,ji2023ddp} and 3D object detection~\citep{chen2020boost,huang2021bevdet,liu2022bevfusion,ge2023metabev}. A genuine 3D geometry representation is crucial for capturing intricate details from 3D annotations, such as road elevations, object heights, and their occlusion patterns, as shown in \Figref{fig:teaser2}. Consequently, generating multi-camera street-view images according to 3D annotations becomes vital to boost downstream perception tasks.

For street-view data synthesis, two pivotal criteria are \textit{realism} and \textit{controllability}. \textit{Realism} requires that the quality of the synthetic data should align with that of real data; and in a given scene, views from varying camera perspectives should remain consistent with one another~\citep{mildenhall2020nerf}.
On the other hand, \textit{controllability} emphasizes the precision in generating street-view images that adhere to provided conditions: the BEV map, 3D object bounding boxes, and camera poses for views. Beyond these core requirements, effective data augmentation should also grant the flexibility to tweak finer scenario attributes, such as prevailing weather conditions or the time of day. Existing solutions like BEVGen~\citep{swerdlow2023street} approach street view generation by encapsulating all semantics within BEV. Conversely, BEVControl~\citep{yang2023bevcontrol} starts by projecting 3D coordinates to image views, subsequently using 2D geometric guidance. However, both methods compromise certain geometric dimensions—height is lost in BEVGen and depth in BEVControl.

The rise of diffusion models has significantly pushed the boundaries of controllable image generation quality. Specifically, ControlNet~\citep{zhang2023adding} proposes a flexible framework to incorporate 2D spatial controls based on pre-trained Text-to-Image (T2I) diffusion models~\citep{rombach2021highresolution}.
However, 3D conditions are distinct from pixel-level conditions or text. The challenge of seamlessly integrating them with multi-camera view consistency in street view synthesis remains.

In this paper, we introduce \methodname, a novel framework dedicated to street-view synthesis with diverse 3D geometry controls\footnote{
In this paper, our 3D geometry controls contain control from road maps, 3D object boxes, and camera poses. We do not consider others like the exact shape of objects or background contents.}.
For \textit{realism}, we harness the power of pre-trained stable diffusion~\citep{rombach2021highresolution}, further fine-tuning it for street view generation.
One distinctive component of our framework is the cross-view attention module. This simple yet effective component provides multi-view consistency through interactions between adjacent views.
In contrast to previous methods, \methodname proposes a separate design for objects and road map encoding to improve \textit{controllability} with 3D data.
More specifically, given the sequence-like, variable-length nature of 3D bounding boxes, we employ cross-attention akin to text embeddings for their encoding.
Besides, we propose that an addictive encoder branch like ControlNet~\citep{zhang2023adding} can encode maps in BEV and is capable of view transformation.
Therefore, our design achieves geometric controls without resorting to any explicit geometric transformations or imposing geometric constraints on multi-camera consistency.
Finally, \methodname factors in textual descriptions, offering attribute control such as weather conditions and time of day.

Our \methodname framework, despite its simplicity, excels in generating strikingly realistic images \& videos that align with road maps, 3D bounding boxes, and varied camera perspectives.
Besides, the images produced can enhance the training for both 3D object detection and BEV segmentation tasks. Furthermore, \methodname offers comprehensive geometric controls at the \textit{scene}, \textit{background}, and \textit{foreground} levels. This flexibility makes it possible to craft previously unseen street views suitable for simulation purposes.
We summarize the main contributions of this work as:
\begin{itemize}[leftmargin=*]
    \item The introduction of \methodname, an innovative framework that generates multi-perspective camera views \& videos conditioned on BEV and 3D data tailored for autonomous driving.
    \item The development of simple yet potent strategies to manage 3D geometric data, effectively addressing the challenges of multi-camera view consistency in street view generation.
    \item Through rigorous experiments, we demonstrate that \methodname outperforms prior street view generation techniques, notably for the multi-dimensional controllability. Additionally, our results reveal that synthetic data delivers considerable improvements in 3D perception tasks. 
\end{itemize}

\section{Related Work}

\textbf{Diffusion Models for Conditional Generation.}
Diffusion models~\citep{ho2020denoising,song2020score,zheng2023non} generate images by learning a progressive denoising process from the Gaussian noise distribution to the image distribution.
These models have proven exceptional across diverse tasks, such as text-to-image synthesis~\citep{rombach2021highresolution,nichol2022glide,yang2023mma}, inpainting~\citep{wang2023imagen}, and instructional image editing~\citep{zhang2023hive, brooks2023instructpix2pix}, due to their adaptability and competence in managing various form of controls~\citep{zhang2023adding,li2023gligen} and multiple conditions~\citep{liu2022compositional,gao2023diffguard}.
Besides, data synthesized from geometric annotations can aid downstream tasks such as 2D object detection~\citep{chen2023integrating,wu2023datasetdm}.
Thus, this paper explores the potential of T2I diffusion models in generating street-view images and benefiting downstream 3D perception models.

\textbf{Street View Generation.}
Numerous street view generation models condition on 2D layouts, such as 2D bounding boxes~\citep{li2023gligen} and semantic segmentation~\citep{wang2022semantic}.
These methods leverage 2D layout information corresponding directly to image scale, whereas the 3D information does not possess this property, thereby rendering such methods unsuitable for leveraging 3D information for generation. 
For street view synthesis with 3D geometry, BEVGen~\citep{swerdlow2023street} is the first to explore.
It utilizes a BEV map as a condition for both roads and vehicles.
However, the omission of height information limits its application in 3D object detection.
BEVControl~\citep{yang2023bevcontrol} amends the loss of object's height by the height-lifting process. Similarly, \cite{wang2023drive} also projects 3D boxes to camera views to guide generation. However, the projection from 3D to 2D results in the loss of essential 3D geometric information, like depth and occlusion.
In this paper, we propose to encode bounding boxes and road maps separately for more nuanced control and integrate scene descriptions, offering enhanced control over the generation of street views.

\textbf{Multi-camera Image Generation} of a 3D scene fundamentally requires viewpoint consistency. Several studies have addressed this issue within the context of indoor scenes. For instance, MVDiffusion~\citep{Tang2023mvdiffusion} employs panoramic images and a cross-view attention module to maintain global consistency, while \cite{poseguideddiffusion} leverage epipolar geometry as a constraining prior. These approaches, however, primarily rely on the continuity of image views, a condition not always met in street views due to limited camera overlap and different camera configurations (\eg, exposure, intrinsic).
Our \methodname introduces extra cross-view attention modules to UNet, which significantly enhances consistency across multi-camera views.
\section{Preliminary}\label{sec:Preliminary}
\noindent\textbf{Problem Formulation.}
In this paper, we consider the coordinate of the LiDAR system as the ego car's coordinate, and parameterize all geometric information according to it.
Let $\mb{S}=\{\mb{M}, \mb{B}, \mb{L}\}$ be the description of a driving scene around the ego vehicle, where $\mb{M}\in\{0, 1\}^{w\times h\times c}$ is the binary map representing a $w\times h$ meter road area in BEV with $c$ semantic classes, $\mb{B}=\{(c_i, b_i)\}_{i=1}^{N}$ represents the 3D bounding box position ($b_i=\{(x_j, y_j, z_j)\}_{j=1}^{8}\in\mathbb{R}^{8\times 3}$) and class ($c_i\in\mathcal{C}$) for each object in the scene, and $\mb{L}$ is the text describing additional information about the scene (\eg, weather and time of day).
Given a camera pose $\mb{P} = [\mb{K}, \mb{R}, \mb{T}]$ (\ie, intrinsics, rotation, and translation), the goal of street-view image generation is to learn a generator $\mathcal{G}(\cdot)$ which synthesizes realistic images $I\in\mathbb{R}^{H\times W\times 3}$ corresponding to the scene $\mb{S}$ and camera pose $\mb{P}$ as, $I=\mathcal{G}(\mb{S}, \mb{P}, z)$, where $z\sim \mathcal{N}(0,1)$ is a random noise from Gaussian distribution.

\noindent\textbf{Conditional Diffusion Models.}
Diffusion models~\citep{ho2020denoising,song2020score} generate data ($\bm{x}_0$) by iteratively denoising a random Gaussian noise ($\bm{x}_{T}$) for $T$ steps.
Typically, to learn the denoising process, the network is trained to predict the noise by minimizing the mean-square error:
\begin{equation}
\ell_{simple} = \mathbb{E}_{\bm{x}_{0},\bm{c},\bm{\epsilon},t}\left[||\bm{\epsilon} - \bm{\epsilon}_{\theta}(\sqrt{\bar{\alpha}_{t}}\bm{x}_{0} + \sqrt{1-\bar{\alpha}_{t}}\bm{\epsilon}, t, \bm{c})||^{2}\right]\text{,}
\label{equ:lsimple}
\end{equation}
where $\bm{\epsilon}_{\theta}$ is the network to train, with parameters $\theta$, $\bm{c}$ is optional conditions, which is used for the conditional generation, $t\in[0,T]$ is the time-step, $\bm{\epsilon}\in\mathcal{N}(0, I)$ is the additive Gaussian noise, and $\bar{\alpha}_{t}$ is a scalar parameter.
Latent diffusion models (LDM)~\citep{rombach2021highresolution} is a special kind of diffusion model, where they utilize a pre-trained Vector Quantized Variational AutoEncoder (VQ-VAE)~\citep{Esser_2021_CVPR} and perform diffusion process in the latent space.
Given the VQ-VAE encoder as $z=\mathcal{E}(x)$, one can rewrite $\bm{\epsilon}_{\theta}(\cdot)$ in \Eqref{equ:lsimple} as $\bm{\epsilon}_{\theta}(\sqrt{\bar{\alpha}_{t}}\mathcal{E}(\bm{x}_{0}) + \sqrt{1-\bar{\alpha}_{t}}\bm{\epsilon}, t, \bm{c})$ for LDM.
Besides, LDM considers text describing the image as condition $c$.
\begin{figure}[t]
    \vspace{-0.3cm}
    \centering
    \includegraphics[width=0.98\linewidth]{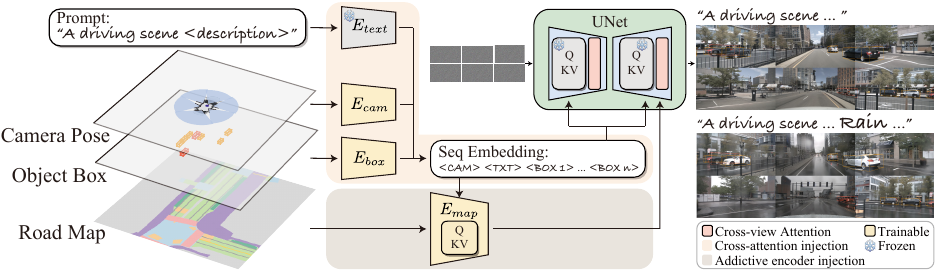}
    \caption{
    Overview of \methodname for street-view image generation.
    \methodname generates highly realistic images, exploiting geometric information from 3D annotations by independently encoding road maps, object boxes, and camera parameters for precise, geometry-guided synthesis. Additionally, \methodname accommodates guidance from descriptive conditions (\eg, weather).
    }
    \label{fig:overview}
    \vspace{-0.2cm}
\end{figure}

\section{Street View Generation with 3D Information}

The overview of \methodname is depicted in \Figref{fig:overview}. Operating on the LDM pipeline, \methodname generates street-view images conditioned on both scene annotations ($\mb{S}$) and the camera pose ($\mb{P}$) for each view. Given the 3D geometric information in scene annotations, projecting all to a BEV map, akin to BEVGen~\citep{swerdlow2023street} or BEVControl~\citep{yang2023bevcontrol}, doesn’t ensure precise guidance for street view generation, as exemplified in \Figref{fig:teaser2}.
Consequently, \methodname categorizes conditions into three levels: \textit{scene} (text and camera pose), \textit{foreground} (3D bounding boxes), and \textit{background} (road map); and integrates them separately via cross-attention and an additive encoder branch, detailed in \Secref{sec:encoding}. Additionally, maintaining consistency across different cameras is crucial for synthesizing street views. Thus, we introduce a simple yet effective cross-view attention module in \Secref{sec:cross-view}. Lastly, we elucidate our training strategies in \Secref{sec:training}, emphasizing Classifier-Free Guidance (CFG) in integrating various conditions.

\subsection{Geometric Conditions Encoding}\label{sec:encoding}

As illustrated in \Figref{fig:overview}, two strategies are employed for information injection into the UNet of diffusion models: cross-attention and additive encoder branch. Given that the attention mechanism~\citep{vaswani2017attention} is tailored for sequential data, cross-attention is apt for managing variable length inputs like text tokens and bounding boxes. Conversely, for grid-like data, such as road maps, the additive encoder branch is effective in information injection~\citep{zhang2023adding}. Therefore, \methodname employs distinct encoding modules for various conditions.

\textbf{Scene-level Encoding} includes camera pose $\mb{P}=\{\mb{K}\in\mathbb{R}^{3\times 3}, \mb{R}\in\mathbb{R}^{3\times3}, \mb{T}\in\mathbb{R}^{3\times 1}\}$, and text sequence $\mb{L}$.
For text, we construct the prompt with a template as 
``{\fontfamily{qcr}\selectfont
A driving scene image at \{location\}. \{description\}
}'',
and leverage a pre-trained CLIP text encoder ($E_{text}$) as LDM~\citep{rombach2021highresolution}, as shown by \Eqref{equ:text_token}, where $L$ is the token length of $\mb{L}$.
As for camera pose, we first concat each parameter by their column, resulting in $\bar{\mb{P}}=[\mb{K}, \mb{R}, \mb{T}]^{T}\in\mathbb{R}^{7\times 3}$.
Since $\bar{\mb{P}}$ contains values from $\sin$/$\cos$ functions and also 3D offsets, to have the model effectively interpret these high-frequency variations, 
we apply Fourier embedding~\citep{mildenhall2020nerf} to each 3-dim vector before leveraging a Multi-Layer Perception (MLP, $E_{cam}$) to embed the camera pose parameters, as in \Eqref{equ:cam_token}.
To maintain consistency, we set the dimension of $h^{c}$ the same as that of ${h^{t}_{i}}$. Through the CLIP text encoder, each text embedding $h^{t}_{i}$ already contains positional information~\citep{radford2021learning}.
Therefore, we prepend the camera pose embedding $h^{c}$ to text embeddings, resulting in scene-level embedding 
$\bm{h}^{s}=[h^{c}, \bm{h}^{t}]$.
\begin{align}
    \bm{h}^{t}&=[h^{t}_{1}\dots h^{t}_{L}]=E_{text}(\mb{L}),
    \label{equ:text_token} \\
    h^{c}&=E_{cam}(\operatorname{Fourier}(\bar{\mb{P}}))=E_{cam}(\operatorname{Fourier}([\mb{K},\mb{R},\mb{T}]^{T}))
    \label{equ:cam_token}.
\end{align}

\textbf{3D Bounding Box Encoding.}
Since each driving scene has a variable length of bounding boxes, we inject them through the cross-attention mechanism similar to scene-level information.
Specifically, we encode each box into a hidden vector $h^{b}$, which has the same dimensions as that of $h^{t}$.
Each 3D bounding box $(c_{i}, b_{i})$ contains two types of information: class label $c_{i}$ and box position $b_{i}$.
For class labels, we utilize the method similar to \cite{li2023gligen}, where the pooled embeddings of class names ($L_{c_{i}}$) are considered as label embeddings.
For box positions $b_{i}\in\mathbb{R}^{8\times 3}$, represented by the coordinates of its 8 corner points, we utilize Fourier embedding to each point and pass through an MLP for encoding, as in \Eqref{equ:box_pos}.
Then, we use an MLP to compress both class and position embedding into one hidden vector, as in \Eqref{equ:box_hidden}.
The final hidden states for all bounding boxes of each scene are represented as $\bm{h}^{b} = [h^{b}_{1}\dots h^{b}_{N}]$, where $N$ is the number of boxes.
\begin{align}
    &e^{b}_{c}(i)=\operatorname{AvgPool}(E_{text}(L_{c_{i}})),\;
    e^{b}_{p}(i)=\operatorname{MLP}_{p}(\operatorname{Fourier}(b_{i})),
    \label{equ:box_pos} \\
    &h^{b}_{i}= E_{box}(c_{i}, b_{i})=\operatorname{MLP}_{b}(e^{b}_{c}(i),e^{b}_{p}(i)).
    \label{equ:box_hidden}
\end{align}
Ideally, the model learns the geometric relationship between bounding boxes and camera pose through training.
However, the distribution of the number of visible boxes to different views is long-tailed.
Thus, we bootstrap learning by filtering visible objects to each view ($v_{i}$), \ie, $f_{viz}$ in \Eqref{equ:box_filter}.
Besides, we also add invisible boxes for augmentation (more details in \Secref{sec:training}).
\begin{equation}
    \bm{h}^{b}_{v_{i}} = \{h^{b}_{i}\in\bm{h}^{b}|f_{viz}(b_{i},\mb{R}_{v_{i}},\mb{T}_{v_{i}}) > 0\}.\label{equ:box_filter}
\end{equation}

\textbf{Road Map Encoding.}
The road map has a 2D-grid format.
While \cite{zhang2023adding} shows the addictive encoder can incorporate this kind of data for 2D guidance,
the inherent perspective differences between the road map's BEV and the camera's First-Person View (FPV) create discrepancies.
BEVControl~\citep{yang2023bevcontrol} employs a back-projection to transform from BEV to FPV but complicates the situation with an ill-posed problem.
In \methodname, we propose that explicit view transformation is unnecessary, as sufficient 3D cues (\eg, height from object boxes and camera pose) allow the addictive encoder to accomplish view transformation.
Specifically, we integrate scene-level and 3D bounding box embeddings into the map encoder (see \Figref{fig:overview}). Scene-level embeddings provide camera poses, and box embeddings offer road elevation cues. Additionally, incorporating text descriptions facilitates the generation of roads under varying conditions (\eg, weather and time of day).
Thus, the map encoder can synergize with other conditions for generation.

\begin{figure}[t]
    \vspace{-0.3cm}
    \centering
    \includegraphics[width=0.98\linewidth]{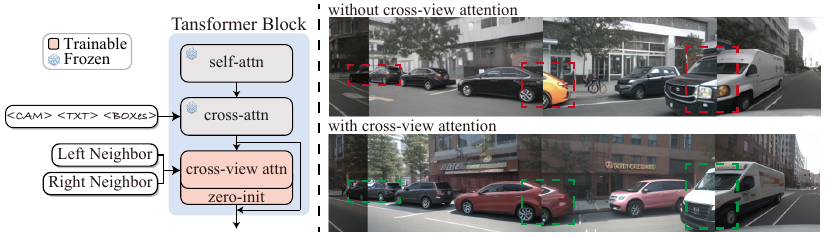}
    \caption{Cross-view Attention.
    \textit{left}: we introduce cross-view attention to the pre-trained UNet after the cross-attention module.
    \textit{right}: we highlight some areas for comparison between without and with cross-view attention. Cross-view attention guarantees consistency across multiple views.
    }
    \label{fig:cross-view_attn}
    \vspace{-0.2cm}
\end{figure}

\subsection{Cross-view Attention Module}\label{sec:cross-view}
In multi-camera view generation, it is crucial that image synthesis remains consistent across different perspectives. To maintain consistency, we introduce a cross-view attention module (\Figref{fig:cross-view_attn}). Given the sparse arrangement of cameras in driving contexts, each cross-view attention allows the target view to access information from its immediate left and right views, as in \Eqref{equ:attn}; here, $t$, $l$, and $r$ are the target, left, and right view respectively. Then, the target view aggregates such information with skip connection, as in \Eqref{equ:skip-attn}, where $\bm{h}^{v}$ indicates the hidden state of the target view.
\begin{align}
    &\operatorname{Attention}_{cv}^{i}(Q_{t}, K_{i}, V_{i}) = \operatorname{softmax}(\begin{matrix}\frac{Q_{t}K_{i}^{T}}{\sqrt{d}}\end{matrix}) \cdot V_{i},\; i\in\{l,r\}\text{,}
    \label{equ:attn} \\
    &\bm{h}_{out}^{v} = \bm{h}_{in}^{v} + \operatorname{Attention}_{cv}^{l} + \operatorname{Attention}_{cv}^{r}\text{.}
    \label{equ:skip-attn}
\end{align}
We inject cross-view attention after the cross-attention module in the UNet and apply zero-initialization~\citep{zhang2023adding} to bootstrap the optimization. 
The efficacy of the cross-view
attention module is demonstrated in \Figref{fig:cross-view_attn} right, \Figref{fig:comparison}, and \Figref{fig:main_results}.
The multilayered structure of UNet enables aggregating information from long-range views after several stacked blocks.
Therefore, using cross-view attention on adjacent views is enough for multi-view consistency, further evidenced by the ablation study in Appendix~\ref{app:ablation_view}.

\subsection{Model Training}\label{sec:training}

\textbf{Classifier-free Guidance} reinforces the impact of conditional guidance~\citep{ho2021classifierfree,rombach2021highresolution}.
For effective CFG, models need to discard conditions during training occasionally.
Given the unique nature of each condition, applying a drop strategy is complex for multiple conditions.
Therefore, our \methodname simplifies this for four conditions by concurrently dropping scene-level conditions (camera pose and text embeddings) at a rate of $\gamma^{s}$. For boxes and maps, which have semantic representations for \textit{null} (i.e., padding token in boxes and $0$ in maps) in their encoding, we maintain them throughout training.
At inference, we utilize \textit{null} for all conditions, enabling meaningful amplification to guide generation.

\textbf{Training Objective and Augmentation.}
With all the conditions injected as inputs, we adapt the training objective described in \Secref{sec:Preliminary} to the multi-condition scenario, as in \Eqref{equ:lfinal}.
\begin{equation}
    \ell = \mathbb{E}_{\bm{x}_{0},\bm{\epsilon},t,\{\bm{S},\bm{P}\}}\left[||\bm{\epsilon} -
    \bm{\epsilon}_{\theta}(\sqrt{\bar{\alpha}_{t}}\mathcal{E}(\bm{x}_{0}) + \sqrt{1-\bar{\alpha}_{t}}\bm{\epsilon}, t, \{\mb{S},\mb{P}\})
    ||\right].
    \label{equ:lfinal}
\end{equation}
Besides, we emphasize two essential strategies when training our \methodname.
First, to counteract our filtering of visible boxes, we randomly add 10\% invisible boxes as an augmentation, enhancing the model's geometric transformation capabilities. Second, to leverage cross-view attention, which facilitates information sharing across multiple views, we apply unique noises to different views in each training step, preventing trivial solutions to \Eqref{equ:lfinal} (\eg, outputting the shared component across different views). Identical random noise is reserved exclusively for inference.

\section{Experiments}
\subsection{Experimental Setups}
\textbf{Dataset and Baselines.}
We employ the nuScenes dataset~\citep{nuScenes}, a prevalent dataset in BEV segmentation and detection for driving, as the testing ground for \methodname. We adhere to the official configuration, utilizing 700 street-view scenes for training and 150 for validation. Our baselines are BEVGen~\citep{swerdlow2023street} and BEVControl~\citep{yang2023bevcontrol}, both recent propositions for street view generation. Our method considers 10 object classes and 8 road classes, surpassing the baseline models in diversity. Appendix~\ref{app:exp_setup} holds additional details.

\textbf{Evaluation Metrics.}
We evaluate both realism and controllability for street view generation.
Realism is mainly measured using Fréchet Inception Distance (FID), reflecting image synthesis quality.
For controllability, \methodname is evaluated through two perception tasks: BEV segmentation and 3D object detection, with CVT~\citep{zhou2022cross} and BEVFusion~\citep{liu2022bevfusion} as perception models, respectively.
Both of them are renowned for their performance in each task.
Firstly, we generate images aligned with the validation set annotations and use perception models pre-trained with real data to assess image quality and control accuracy.
Then, data is generated based on the training set to examine the support for training perception models as data augmentation.

\textbf{Model Setup.}
Our \methodname utilizes pre-trained weights from Stable Diffusion v1.5, training only newly added parameters. Per~\cite{zhang2023adding}, a trainable UNet encoder is created for $E_{map}$. New parameters, except for the zero-init module and the class token, are randomly initialized. We adopt two resolutions to reconcile discrepancies in perception tasks and baselines: 224$\times$400 (0.25$\times$ down-sample) following BEVGen and for CVT model support, and a higher 272$\times$736 (0.5$\times$ down-sample) for BEVFusion support. Unless stated otherwise, images are sampled using the UniPC~\citep{zhao2023unipc} scheduler for 20 steps with CFG at $2.0$.

\begin{table}[t]
\vspace{-0.3cm}
\centering
\caption{Comparison of generation fidelity with driving-view generation methods. 
Conditions for data synthesis are from nuScenes validation set. For each task, we test the corresponding models trained on the nuScenes training set. \methodname surpasses all baselines throughout the evaluation. $\uparrow$/$\downarrow$ indicates that a higher/lower value is better. The best results are in \textbf{bold}, while the second best results are in \uit{underlined italic} (when other methods are available).}
\label{tab:test}
\resizebox{0.95\textwidth}{!}{
\begin{tabular}{l|c|c|c|c|c|c}
\toprule
\multirow{2}[3]{*}{Method} &
  \multirow{2}[3]{*}{\begin{tabular}[c]{@{}c@{}}Synthesis\\ resolution\end{tabular}} &
  \multirow{2}[3]{*}{FID$\downarrow$} &
  \multicolumn{2}{c|}{BEV segmentation} &
  \multicolumn{2}{c}{3D object detection} \\
  \cmidrule{4-7} 
  &&&
  Road mIoU $\uparrow$ &
  Vehicle mIoU $\uparrow$ &
  mAP $\uparrow$ &
  NDS $\uparrow$   \\
\midrule
Oracle
    & -
    & -
    & 72.21
    & 33.66
    & 35.54
    & 41.21 \\
Oracle
    & 224$\times$400
    & -
    & 72.19
    & 33.61
    & 23.54
    & 31.08 \\
\midrule
BEVGen
    & 224$\times$400
    & 25.54
    & 50.20
    & 5.89
    & -
    & -     \\
BEVControl
    & -
    & 24.85
    & \uit{60.80}
    & 26.80
    & -
    & -     \\ 
\midrule
\methodname
    & 224$\times$400
    & \textbf{16.20}
    & \textbf{61.05}
    & \uit{27.01}
    & 12.30
    & 23.32 \\
\methodname
    & 272$\times$736
    & \uit{16.59}
    & 54.24
    & \textbf{31.05}
    & \textbf{20.85}
    & \textbf{30.26} \\
\bottomrule
\end{tabular}
}%
\vspace{-0.3cm}
\end{table}
\begin{figure}[t]
    \centering
    \includegraphics[width=0.98\linewidth]{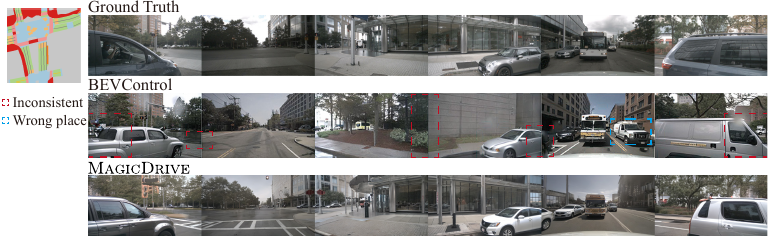}
    \vspace{-0.3cm}
    \caption{Qualitative comparison with BEVControl on driving scene from nuScenes validation set. We highlight some areas with rectangles to ease comparison. Compared with BEVControl, generations from \methodname appear more consistent in both background and foreground. 
    }
    \label{fig:comparison}
    \vspace{-0.5cm}
\end{figure}

\begin{figure}[t]
\vspace{-0.3cm}
\begin{minipage}{\textwidth}
    \begin{minipage}[t][][b]{0.55\textwidth}
        \centering
        \captionsetup{width=.95\linewidth}
        \captionof{table}{
        Comparison about support for 3D object detection model (\ie, BEVFusion). \methodname generates 272×736 images for augmentation. Results are reported on the nuScenes validation set.
        }
        \label{tab:bevfusion_train}
        \vspace{0.03cm}
        \resizebox{0.95\textwidth}{!}{
        \begin{tabular}{cc|c|c}
        \toprule
        Modality & Data & mAP $\uparrow$ & NDS $\uparrow$ \\
        \midrule
        \multirow{2}{*}{C}
           & w/o synthetic data & 32.88 & 37.81 \\
           & w/ \methodname
           & \textbf{35.40 {\tiny\textcolor{green}{+2.52}}}
           & \textbf{39.76 {\tiny\textcolor{green}{+1.95}}}
           \\
        \midrule
        \multirow{2}{*}{C+L} 
           & w/o synthetic data &65.40 & 69.59 \\
           & w/ \methodname
           & \textbf{67.86 {\tiny\textcolor{green}{+2.46}}}
           & \textbf{70.72 {\tiny\textcolor{green}{+1.13}}}
           \\
        \bottomrule
        \end{tabular}
        }%
    \end{minipage}
    \hfill
    \begin{minipage}[t][][b]{0.43\textwidth}
        \centering
        \captionof{table}{
        Comparison about support for BEV segmentation model (\ie, CVT). Results are reported by testing on the nuScenes validation set.
        }
        \label{tab:cvt_train}
        \resizebox{0.95\textwidth}{!}{
        \begin{tabular}{l|c|c@{}}
        \toprule
        \multirow{2}{*}{Data} & 
        \multirow{2}{*}{\begin{tabular}[c]{@{}c@{}}Vehicle\\ mIoU $\uparrow$\end{tabular}} &
        \multirow{2}{*}{\begin{tabular}[c]{@{}c@{}}Road\\ mIoU $\uparrow$\end{tabular}} \\
        &&\\
        
        \midrule
        w/o synthetic data & 36.00 & 74.30 \\
        \midrule
        w/ BEVGen & 36.60 {\tiny\textcolor{green}{+0.60}} & 71.90 {\tiny\textcolor{red}{-2.40}} \\
        w/ \methodname & \textbf{40.34 {\tiny\textcolor{green}{+4.34}}} & \textbf{79.56 {\tiny\textcolor{green}{+5.26}}} \\
        \bottomrule
        \end{tabular}
        }%
        
    \end{minipage}
    \vspace{0.2cm}
\end{minipage}
\begin{minipage}{\textwidth}
    \centering
    \includegraphics[width=0.98\linewidth]{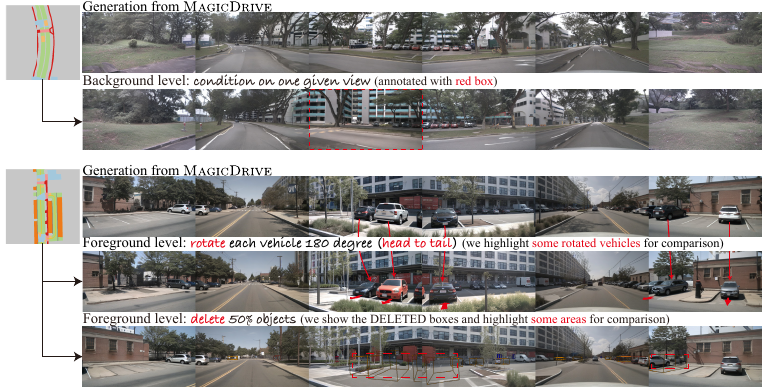}
    \vspace{-0.3cm}
    \captionof{figure}{Showcase for multi-level control with \methodname. We show background-level and foreground-level control separately with different conditions. All scenes are based on the nuScenes validation set. More results are in Appendix~\ref{sec:scene-level}.}
    \label{fig:main_results}
\end{minipage}
\vspace{-0.5cm}
\end{figure}

\subsection{Main Results}

\textbf{Realism and Controllability Validation.}
We assess \methodname's capability to create realistic street-view images with the annotations from the nuScenes validation set. As shown by \Tabref{tab:test}, \methodname outperforms others in image quality, yielding notably lower FID scores. 
Regarding controllability, assessed via BEV segmentation tasks, \methodname equals or exceeds baseline results at 224$\times$400 resolution due to the distinct encoding design that enhances vehicle generation precision.
At 272$\times$736 resolution, our encoding strategy advancements enhance vehicle mIoU performance. Cropping large areas negatively impacts road mIoU on CVT. However, our bounding box encoding efficacy is backed by BEVFusion's results in 3D object detection. 

\textbf{Training Support for BEV Segmentation and 3D Object Detection.}
\methodname can produce augmented data with accurate annotation controls, enhancing the training for perception tasks.
For BEV segmentation, we augment an equal number of images as in the original dataset, ensuring consistent training iterations and batch sizes for fair comparisons to the baseline.
As shown in \Tabref{tab:cvt_train}, \methodname significantly enhances CVT in both settings, outperforming BEVGen, which only marginally improves vehicle segmentation.
For 3D object detection, we train BEVFusion models with \methodname’s synthetic data as augmentation.
To optimize data augmentation, we randomly exclude 50\% of bounding boxes in each generated scene.
\Tabref{tab:bevfusion_train} shows the advantageous impact of \methodname’s data in both CAM-only (C) and CAM+LiDAR (C+L) settings.
It’s crucial to note that in CAM+LiDAR settings, BEVFusion utilizes both modalities for object detection,
requiring more precise image generation due to LiDAR data incorporation.
Nevertheless, \methodname's synthetic data integrates seamlessly with LiDAR inputs, highlighting the data’s high fidelity.

\subsection{Qualitative Evaluation}
\textbf{Comparison with Baselines.}
We assessed \methodname against two baselines, BEVGen and BEVControl, synthesizing multi-camera views for the same validation scenes (the comparison with BEVGen is in the Appendix \ref{sec:comp-bevgen}).
\Figref{fig:comparison} illustrates that \methodname generates images markedly superior in quality to BEVControl, particularly excelling in accurate object positioning and maintaining consistency in street views for backgrounds and objects.
Such performance primarily stems from \methodname’s
bounding box encoder and its cross-view attention module.

\textbf{Multi-level Controls.}
The design of \methodname introduces multi-level controls to street-view generation through separation encoding. This section demonstrates the capabilities of \methodname by exploring three control signal levels: \textit{scene level} (time of day and weather), \textit{background level} (BEV map alterations and conditional views), and \textit{foreground level} (object orientation and deletion). As illustrated in \Figref{fig:teaser}, \Figref{fig:main_results}, and Appendix~\ref{sec:scene-level}, \methodname adeptly accommodates alterations at each level, maintaining multi-camera consistency and high realism in generation.

\subsection{Extension to Video Generation}
\begin{figure}[h]
    \centering
    \includegraphics[width=\linewidth]{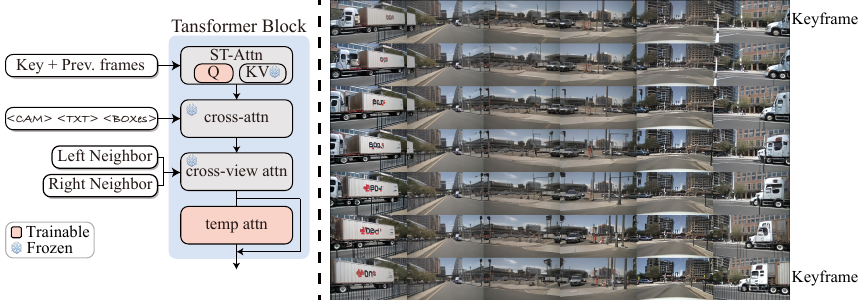}
    \caption{Extend \methodname to video generation. \textit{Left}: changes in transformer block for video generation. \textit{Right}: result of video generation. Only keyframes have bounding box control.}
    \label{fig:temporal}
\end{figure}

We demonstrate the extensibility of \methodname to video generation by fine-tuning it on nuScenes videos. This involves modifying self-attention to ST-Attn~\citep{wu2023tune}, adding a temporal attention module to each transformer block (\Figref{fig:temporal} \textit{left}), and tuning the model on 7-frame clips with only the first and the last frames having bounding boxes. We sample initial noise independently for each frame using the UniPC~\citep{zhao2023unipc} sampler for 20 steps and illustrate an example in \Figref{fig:temporal} \textit{right}. 

Furthermore, by utilizing the interpolated annotations from ASAP~\citep{wang2023we} like DriveDreamer~\citep{wang2023drive}, MagicDrive can be extended to 16-frame video generation at 12Hz trained on Nvidia V100 GPUs. More results (\textit{e.g.}, video visualization) can be found on our website.

\section{Ablation Study}\label{sec:ablation}

\begin{figure}[t]
\vspace{-0.3cm}
\begin{minipage}{0.98\textwidth}
    \begin{minipage}[t][][b]{0.6\textwidth}
        \vspace{-0.3cm}
        \captionof{table}{Ablation of the separate box encoder. Evaluation results are from CVT on the synthetic nuScenes validation set, without $M=\{0\}$ in CFG scale $=2$. \methodname has better controllability and keeps image quality.}
        \label{tab:box_ablation}
        \begin{center}
        \begin{tabular}[b]{@{}l|ccc@{}}
        \toprule
        \multirow{2}{*}{Method} 
        & \multirow{2}{*}{FID $\downarrow$}
        & \multirow{2}{*}{\begin{tabular}[c]{@{}c@{}}Road\\ mIoU $\uparrow$\end{tabular}}
        & \multirow{2}{*}{\begin{tabular}[c]{@{}c@{}}Vehicle\\ mIoU $\uparrow$\end{tabular}} \\
        &&&\\

        \midrule
        w/o $E_{box}$ & 18.06 & 58.31 & 5.50 \\
        w/o $f_{viz}$ & 14.67 & 56.46 & 24.73 \\
        w/ $E_{box}$ \& map$_{obj}$ & 14.70 & 56.04 & 26.20 \\
        \midrule
        Ours
            & \textbf{14.46}
            & \textbf{59.31}
            & \textbf{27.13} \\
        \bottomrule
        \end{tabular}
        \end{center}
    \end{minipage}
    \hfill
    \begin{minipage}[t][][b]{0.38\textwidth}
        \centering
        \includegraphics[width=0.98\textwidth]{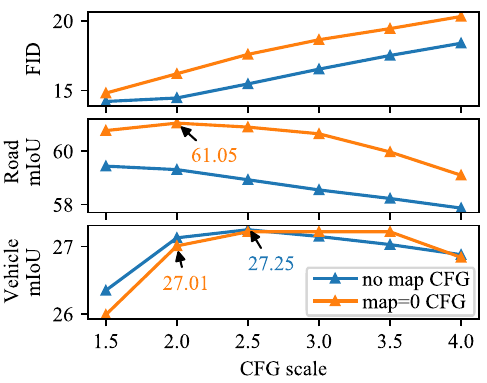}
        \vspace{-0.3cm}
        \captionof{figure}{Effect of CFG on different conditions to each metrics.}\label{fig:ablation_cfg}
    \end{minipage}
\end{minipage}
\vspace{-0.5cm}
\end{figure}

\textbf{Bounding Box Encoding.}
\methodname utilizes separate encoders for bounding boxes and road maps.
To demonstrate the efficacy, we train a ControlNet~\citep{zhang2023adding} that takes the BEV map with both road and object semantics as a condition (like BEVGen), denoted as ``w/o $E_{box}$'' in \Tabref{tab:box_ablation}.
Objects in BEV maps are relatively small, which require separate $E_{box}$ for accurate vehicle annotations, as shown by the vehicle mIoU performance gap. Applying visible object filter $f_{viz}$ significantly improves both road and vehicle mIoU by reducing the optimization burden. A \methodname variant incorporating $E_{box}$ with BEV of road and object semantics didn't enhance performance, emphasizing the importance of integrating diverse information through different strategies. 

\textbf{Effect of Classifier-free Guidance.}
We focus on the two most crucial conditions, \ie object boxes and road maps, and analyze how CFG affects the performance of generation.
We change CFG from 1.5 to 4.0 and plot the change of validation results from CVT in \Figref{fig:ablation_cfg}.
Firstly, by increasing CFG scale, FID degrades due to notable changes in contrast and sharpness, as seen in previous studies~\citep{chen2023integrating}.
Secondly, retaining the same map for both conditional and unconditional inference eliminates CFG’s effect on the map condition.
As shown by blue lines of \Figref{fig:ablation_cfg}, increasing CFG scale results in the highest vehicle mIoU at CFG=2.5, but the road mIoU keeps decreasing.
Thirdly, with $M=\{0\}$ for unconditional inference in CFG, road mIoU significantly increases.
However, it slightly degrades the guidance on vehicle generation.
As mentioned in \Secref{sec:training}, CFG complexity increases with more conditions. Despite simplifying training, various CFG choices exist during inference. 
We leave the in-depth investigation for this case as future work.
\section{Conclusion}
This paper presents \methodname, a novel framework to encode multiple geometric controls for high-quality multi-camera street view generation.
With the separation encoding design, \methodname fully utilizes geometric information from 3D annotations and realizes accurate semantic control for street views.
Besides, the proposed cross-view attention module is simple yet effective in guaranteeing consistency across multi-camera views.
As evidenced by experiments, the generations from \methodname show high realism and fidelity to 3D annotations.
Multiple controls equipped \methodname with improved generalizability for the generation of novel street views.
Meanwhile, \methodname can be used for data augmentation, facilitating the training for perception models on both BEV segmentation and 3D object detection tasks.

\begin{wrapfigure}{r}{6.5cm}
\vspace{-0.5cm}
\includegraphics[width=0.98\linewidth]{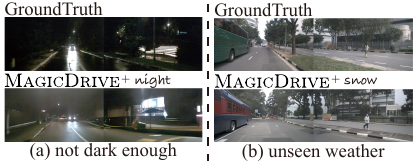}
\vspace{-0.2cm}
\caption{Failure cases of \methodname.}\label{fig:limitations}
\vspace{-0.5cm}
\end{wrapfigure}
\textbf{Limitation and Future Work.}
We show failure cases from \methodname in \Figref{fig:limitations}.
Although \methodname can generate night views, they are not as dark as real images (as in \Figref{fig:limitations}a).
This may be due to that diffusion models are hard to generate too dark images~\citep{crosslabsDiffusionWith}.
\Figref{fig:limitations}b shows that \methodname cannot generate unseen weathers for nuScenes.
Future work may focus on how to improve the cross-domain generalization ability of street view generation.

\textbf{Acknowledgement.}
This work is supported in part by the General Research Fund (GRF) of Hong Kong Research Grants Council (RGC) under Grant No. 14203521, in part by the CUHK SSFCRS funding No. 3136023, and in part by the Research Matching Grant Scheme under Grant No. 7106937, 8601130, and 8601440.
We gratefully acknowledge the support of MindSpore, CANN (Compute Architecture for Neural Networks)
and Ascend AI Processor used for this research.
This research has been made possible by funding support from the Research Grants Council of Hong Kong through the Research Impact Fund project R6003-21.

\bibliography{main}

\begin{thebibliography}{51}
\providecommand{\natexlab}[1]{#1}
\providecommand{\url}[1]{\texttt{#1}}
\expandafter\ifx\csname urlstyle\endcsname\relax
  \providecommand{\doi}[1]{doi: #1}\else
  \providecommand{\doi}{doi: \begingroup \urlstyle{rm}\Url}\fi

\bibitem[Brooks et~al.(2023)Brooks, Holynski, and Efros]{brooks2023instructpix2pix}
Tim Brooks, Aleksander Holynski, and Alexei~A Efros.
\newblock Instructpix2pix: Learning to follow image editing instructions.
\newblock In \emph{CVPR}, 2023.

\bibitem[Caesar et~al.(2020)Caesar, Bankiti, Lang, Vora, Liong, Xu, Krishnan, Pan, Baldan, and Beijbom]{nuScenes}
Holger Caesar, Varun Bankiti, Alex~H Lang, Sourabh Vora, Venice~Erin Liong, Qiang Xu, Anush Krishnan, Yu~Pan, Giancarlo Baldan, and Oscar Beijbom.
\newblock nuscenes: A multimodal dataset for autonomous driving.
\newblock In \emph{CVPR}, 2020.

\bibitem[Chen et~al.(2021)Chen, Hong, Xu, Li, and Yeung]{chen2021multisiam}
Kai Chen, Lanqing Hong, Hang Xu, Zhenguo Li, and Dit-Yan Yeung.
\newblock Multisiam: Self-supervised multi-instance siamese representation learning for autonomous driving.
\newblock In \emph{ICCV}, 2021.

\bibitem[Chen et~al.(2023{\natexlab{a}})Chen, Liu, Hong, Xu, Li, and Yeung]{chen2023mixed}
Kai Chen, Zhili Liu, Lanqing Hong, Hang Xu, Zhenguo Li, and Dit-Yan Yeung.
\newblock Mixed autoencoder for self-supervised visual representation learning.
\newblock In \emph{CVPR}, 2023{\natexlab{a}}.

\bibitem[Chen et~al.(2023{\natexlab{b}})Chen, Wang, Yang, Han, Hong, Mi, Xu, Liu, Huang, Li, Yeung, Shang, Jiang, and Liu]{chen2023gaining}
Kai Chen, Chunwei Wang, Kuo Yang, Jianhua Han, Lanqing Hong, Fei Mi, Hang Xu, Zhengying Liu, Wenyong Huang, Zhenguo Li, Dit-Yan Yeung, Lifeng Shang, Xin Jiang, and Qun Liu.
\newblock Gaining wisdom from setbacks: Aligning large language models via mistake analysis.
\newblock \emph{arXiv preprint arXiv:2310.10477}, 2023{\natexlab{b}}.

\bibitem[Chen et~al.(2023{\natexlab{c}})Chen, Xie, Chen, Hong, Li, and Yeung]{chen2023integrating}
Kai Chen, Enze Xie, Zhe Chen, Lanqing Hong, Zhenguo Li, and Dit-Yan Yeung.
\newblock Integrating geometric control into text-to-image diffusion models for high-quality detection data generation via text prompt.
\newblock \emph{arXiv preprint arXiv:2306.04607}, 2023{\natexlab{c}}.

\bibitem[Chen et~al.(2020)Chen, Li, Gao, and Zhao]{chen2020boost}
Yaran Chen, Haoran Li, Ruiyuan Gao, and Dongbin Zhao.
\newblock Boost 3-d object detection via point clouds segmentation and fused 3-d giou-l1 loss.
\newblock \emph{IEEE TNNLS}, 2020.

\bibitem[Esser et~al.(2021)Esser, Rombach, and Ommer]{Esser_2021_CVPR}
Patrick Esser, Robin Rombach, and Bjorn Ommer.
\newblock Taming transformers for high-resolution image synthesis.
\newblock In \emph{CVPR}, 2021.

\bibitem[Gao et~al.(2023)Gao, Zhao, Hong, and Xu]{gao2023diffguard}
Ruiyuan Gao, Chenchen Zhao, Lanqing Hong, and Qiang Xu.
\newblock {DiffGuard}: Semantic mismatch-guided out-of-distribution detection using pre-trained diffusion models.
\newblock In \emph{ICCV}, 2023.

\bibitem[Ge et~al.(2023)Ge, Chen, Xie, Wang, Hong, Lu, Li, and Luo]{ge2023metabev}
Chongjian Ge, Junsong Chen, Enze Xie, Zhongdao Wang, Lanqing Hong, Huchuan Lu, Zhenguo Li, and Ping Luo.
\newblock {MetaBEV}: Solving sensor failures for bev detection and map segmentation.
\newblock In \emph{ICCV}, 2023.

\bibitem[Gou et~al.(2023)Gou, Liu, Chen, Hong, Xu, Li, Yeung, Kwok, and Zhang]{gou2023mixture}
Yunhao Gou, Zhili Liu, Kai Chen, Lanqing Hong, Hang Xu, Aoxue Li, Dit-Yan Yeung, James~T Kwok, and Yu~Zhang.
\newblock Mixture of cluster-conditional lora experts for vision-language instruction tuning.
\newblock \emph{arXiv preprint arXiv:2312.12379}, 2023.

\bibitem[Guttenberg(2023)]{crosslabsDiffusionWith}
Nicholas Guttenberg.
\newblock Diffusion with offset noise.
\newblock \url{https://www.crosslabs.org/blog/diffusion-with-offset-noise}, 2023.

\bibitem[Han et~al.(2021)Han, Liang, Xu, Chen, Hong, Ye, Zhang, Li, Liang, and Xu]{han2021soda10m}
Jianhua Han, Xiwen Liang, Hang Xu, Kai Chen, Lanqing Hong, Chaoqiang Ye, Wei Zhang, Zhenguo Li, Xiaodan Liang, and Chunjing Xu.
\newblock Soda10m: Towards large-scale object detection benchmark for autonomous driving.
\newblock \emph{arXiv preprint arXiv:2106.11118}, 2021.

\bibitem[Ho \& Salimans(2021)Ho and Salimans]{ho2021classifierfree}
Jonathan Ho and Tim Salimans.
\newblock Classifier-free diffusion guidance.
\newblock In \emph{NeurIPS 2021 Workshop on Deep Generative Models and Downstream Applications}, 2021.

\bibitem[Ho et~al.(2020)Ho, Jain, and Abbeel]{ho2020denoising}
Jonathan Ho, Ajay Jain, and Pieter Abbeel.
\newblock Denoising diffusion probabilistic models.
\newblock In \emph{NeurIPS}, 2020.

\bibitem[Huang et~al.(2021)Huang, Huang, Zhu, Yun, and Du]{huang2021bevdet}
Junjie Huang, Guan Huang, Zheng Zhu, Ye~Yun, and Dalong Du.
\newblock Bevdet: High-performance multi-camera 3d object detection in bird-eye-view.
\newblock \emph{arXiv preprint arXiv:2112.11790}, 2021.

\bibitem[Ji et~al.(2023)Ji, Chen, Xie, Hong, Liu, Liu, Lu, Li, and Luo]{ji2023ddp}
Yuanfeng Ji, Zhe Chen, Enze~Xie Xie, Lanqing Hong, Xihui Liu, Zhaoqiang Liu, Tong Lu, Zhenguo Li, and Ping Luo.
\newblock {DDP}: Diffusion model for dense visual prediction.
\newblock In \emph{ICCV}, 2023.

\bibitem[Li et~al.(2022)Li, Chen, Wang, Hong, Ye, Han, Chen, Zhang, Xu, Yeung, et~al.]{li2022coda}
Kaican Li, Kai Chen, Haoyu Wang, Lanqing Hong, Chaoqiang Ye, Jianhua Han, Yukuai Chen, Wei Zhang, Chunjing Xu, Dit-Yan Yeung, et~al.
\newblock Coda: A real-world road corner case dataset for object detection in autonomous driving.
\newblock \emph{arXiv preprint arXiv:2203.07724}, 2022.

\bibitem[Li et~al.(2023{\natexlab{a}})Li, Liu, Chen, Hong, Zhuge, Yeung, Lu, and Jia]{li2023trackdiffusion}
Pengxiang Li, Zhili Liu, Kai Chen, Lanqing Hong, Yunzhi Zhuge, Dit-Yan Yeung, Huchuan Lu, and Xu~Jia.
\newblock Trackdiffusion: Multi-object tracking data generation via diffusion models.
\newblock \emph{arXiv preprint arXiv:2312.00651}, 2023{\natexlab{a}}.

\bibitem[Li et~al.(2023{\natexlab{b}})Li, Liu, Wu, Mu, Yang, Gao, Li, and Lee]{li2023gligen}
Yuheng Li, Haotian Liu, Qingyang Wu, Fangzhou Mu, Jianwei Yang, Jianfeng Gao, Chunyuan Li, and Yong~Jae Lee.
\newblock Gligen: Open-set grounded text-to-image generation.
\newblock In \emph{CVPR}, 2023{\natexlab{b}}.

\bibitem[Lin et~al.(2014)Lin, Maire, Belongie, Hays, Perona, Ramanan, Doll{\'a}r, and Zitnick]{lin2014microsoft}
Tsung-Yi Lin, Michael Maire, Serge Belongie, James Hays, Pietro Perona, Deva Ramanan, Piotr Doll{\'a}r, and C~Lawrence Zitnick.
\newblock Microsoft coco: Common objects in context.
\newblock In \emph{ECCV}, 2014.

\bibitem[Liu et~al.(2022{\natexlab{a}})Liu, Li, Du, Torralba, and Tenenbaum]{liu2022compositional}
Nan Liu, Shuang Li, Yilun Du, Antonio Torralba, and Joshua~B Tenenbaum.
\newblock Compositional visual generation with composable diffusion models.
\newblock In \emph{ECCV}, 2022{\natexlab{a}}.

\bibitem[Liu et~al.(2021)Liu, Lin, Cao, Hu, Wei, Zhang, Lin, and Guo]{liu2021Swin}
Ze~Liu, Yutong Lin, Yue Cao, Han Hu, Yixuan Wei, Zheng Zhang, Stephen Lin, and Baining Guo.
\newblock Swin transformer: Hierarchical vision transformer using shifted windows.
\newblock In \emph{ICCV}, 2021.

\bibitem[Liu et~al.(2023{\natexlab{a}})Liu, Tang, Amini, Yang, Mao, Rus, and Han]{liu2022bevfusion}
Zhijian Liu, Haotian Tang, Alexander Amini, Xingyu Yang, Huizi Mao, Daniela Rus, and Song Han.
\newblock Bevfusion: Multi-task multi-sensor fusion with unified bird's-eye view representation.
\newblock In \emph{ICRA}, 2023{\natexlab{a}}.

\bibitem[Liu et~al.(2022{\natexlab{b}})Liu, Han, Chen, Hong, Xu, Xu, and Li]{liu2022task}
Zhili Liu, Jianhua Han, Kai Chen, Lanqing Hong, Hang Xu, Chunjing Xu, and Zhenguo Li.
\newblock Task-customized self-supervised pre-training with scalable dynamic routing.
\newblock In \emph{AAAI}, 2022{\natexlab{b}}.

\bibitem[Liu et~al.(2023{\natexlab{b}})Liu, Chen, Zhang, Han, Hong, Xu, Li, Yeung, and Kwok]{liu2023geomerasing}
Zhili Liu, Kai Chen, Yifan Zhang, Jianhua Han, Lanqing Hong, Hang Xu, Zhenguo Li, Dit-Yan Yeung, and James Kwok.
\newblock Geom-erasing: Geometry-driven removal of implicit concept in diffusion models.
\newblock \emph{arXiv preprint arXiv:2310.05873}, 2023{\natexlab{b}}.

\bibitem[Loshchilov \& Hutter(2019)Loshchilov and Hutter]{loshchilov2018decoupled}
Ilya Loshchilov and Frank Hutter.
\newblock Decoupled weight decay regularization.
\newblock In \emph{ICLR}, 2019.

\bibitem[Mildenhall et~al.(2020)Mildenhall, Srinivasan, Tancik, Barron, Ramamoorthi, and Ng]{mildenhall2020nerf}
Ben Mildenhall, Pratul~P Srinivasan, Matthew Tancik, Jonathan~T Barron, Ravi Ramamoorthi, and Ren Ng.
\newblock Nerf: Representing scenes as neural radiance fields for view synthesis.
\newblock In \emph{ECCV}, 2020.

\bibitem[Nichol et~al.(2022)Nichol, Dhariwal, Ramesh, Shyam, Mishkin, Mcgrew, Sutskever, and Chen]{nichol2022glide}
Alexander~Quinn Nichol, Prafulla Dhariwal, Aditya Ramesh, Pranav Shyam, Pamela Mishkin, Bob Mcgrew, Ilya Sutskever, and Mark Chen.
\newblock Glide: Towards photorealistic image generation and editing with text-guided diffusion models.
\newblock In \emph{ICML}, 2022.

\bibitem[Radford et~al.(2021)Radford, Kim, Hallacy, Ramesh, Goh, Agarwal, Sastry, Askell, Mishkin, Clark, et~al.]{radford2021learning}
Alec Radford, Jong~Wook Kim, Chris Hallacy, Aditya Ramesh, Gabriel Goh, Sandhini Agarwal, Girish Sastry, Amanda Askell, Pamela Mishkin, Jack Clark, et~al.
\newblock Learning transferable visual models from natural language supervision.
\newblock In \emph{ICML}, 2021.

\bibitem[Rombach et~al.(2022)Rombach, Blattmann, Lorenz, Esser, and Ommer]{rombach2021highresolution}
Robin Rombach, Andreas Blattmann, Dominik Lorenz, Patrick Esser, and Bj{\"o}rn Ommer.
\newblock High-resolution image synthesis with latent diffusion models.
\newblock In \emph{CVPR}, 2022.

\bibitem[Song et~al.(2020)Song, Sohl-Dickstein, Kingma, Kumar, Ermon, and Poole]{song2020score}
Yang Song, Jascha Sohl-Dickstein, Diederik~P Kingma, Abhishek Kumar, Stefano Ermon, and Ben Poole.
\newblock Score-based generative modeling through stochastic differential equations.
\newblock In \emph{ICLR}, 2020.

\bibitem[Swerdlow et~al.(2023)Swerdlow, Xu, and Zhou]{swerdlow2023street}
Alexander Swerdlow, Runsheng Xu, and Bolei Zhou.
\newblock Street-view image generation from a bird's-eye view layout.
\newblock \emph{arXiv preprint arXiv:2301.04634}, 2023.

\bibitem[Tang et~al.(2023)Tang, Zhang, Chen, Wang, and Furukawa]{Tang2023mvdiffusion}
Shitao Tang, Fuyang Zhang, Jiacheng Chen, Peng Wang, and Yasutaka Furukawa.
\newblock Mvdiffusion: Enabling holistic multi-view image generation with correspondence-aware diffusion.
\newblock \emph{arXiv preprint arXiv:2307.01097}, 2023.

\bibitem[Tseng et~al.(2023)Tseng, Li, Kim, Alsisan, Huang, and Kopf]{poseguideddiffusion}
Hung-Yu Tseng, Qinbo Li, Changil Kim, Suhib Alsisan, Jia-Bin Huang, and Johannes Kopf.
\newblock Consistent view synthesis with pose-guided diffusion models.
\newblock In \emph{CVPR}, 2023.

\bibitem[Vaswani et~al.(2017)Vaswani, Shazeer, Parmar, Uszkoreit, Jones, Gomez, Kaiser, and Polosukhin]{vaswani2017attention}
Ashish Vaswani, Noam Shazeer, Niki Parmar, Jakob Uszkoreit, Llion Jones, Aidan~N Gomez, {\L}ukasz Kaiser, and Illia Polosukhin.
\newblock Attention is all you need.
\newblock In \emph{NeurIPS}, 2017.

\bibitem[Wang et~al.(2023{\natexlab{a}})Wang, Saharia, Montgomery, Pont-Tuset, Noy, Pellegrini, Onoe, Laszlo, Fleet, Soricut, et~al.]{wang2023imagen}
Su~Wang, Chitwan Saharia, Ceslee Montgomery, Jordi Pont-Tuset, Shai Noy, Stefano Pellegrini, Yasumasa Onoe, Sarah Laszlo, David~J Fleet, Radu Soricut, et~al.
\newblock Imagen editor and editbench: Advancing and evaluating text-guided image inpainting.
\newblock In \emph{CVPR}, 2023{\natexlab{a}}.

\bibitem[Wang et~al.(2022)Wang, Bao, Zhou, Chen, Chen, Yuan, and Li]{wang2022semantic}
Weilun Wang, Jianmin Bao, Wengang Zhou, Dongdong Chen, Dong Chen, Lu~Yuan, and Houqiang Li.
\newblock Semantic image synthesis via diffusion models.
\newblock \emph{arXiv preprint arXiv:2207.00050}, 2022.

\bibitem[Wang et~al.(2023{\natexlab{b}})Wang, Zhu, Huang, Chen, Zhu, and Lu]{wang2023drive}
Xiaofeng Wang, Zheng Zhu, Guan Huang, Xinze Chen, Jiagang Zhu, and Jiwen Lu.
\newblock Drivedreamer: Towards real-world-driven world models for autonomous driving.
\newblock \emph{arXiv preprint arXiv:2309.09777}, 2023{\natexlab{b}}.

\bibitem[Wang et~al.(2023{\natexlab{c}})Wang, Zhu, Zhang, Huang, Ye, Xu, Chen, and Wang]{wang2023we}
Xiaofeng Wang, Zheng Zhu, Yunpeng Zhang, Guan Huang, Yun Ye, Wenbo Xu, Ziwei Chen, and Xingang Wang.
\newblock Are we ready for vision-centric driving streaming perception? the asap benchmark.
\newblock In \emph{CVPR}, 2023{\natexlab{c}}.

\bibitem[Wu et~al.(2023{\natexlab{a}})Wu, Ge, Wang, Lei, Gu, Shi, Hsu, Shan, Qie, and Shou]{wu2023tune}
Jay~Zhangjie Wu, Yixiao Ge, Xintao Wang, Stan~Weixian Lei, Yuchao Gu, Yufei Shi, Wynne Hsu, Ying Shan, Xiaohu Qie, and Mike~Zheng Shou.
\newblock Tune-a-video: One-shot tuning of image diffusion models for text-to-video generation.
\newblock In \emph{CVPR}, 2023{\natexlab{a}}.

\bibitem[Wu et~al.(2023{\natexlab{b}})Wu, Zhao, Chen, Gu, Zhao, He, Zhou, Shou, and Shen]{wu2023datasetdm}
Weijia Wu, Yuzhong Zhao, Hao Chen, Yuchao Gu, Rui Zhao, Yefei He, Hong Zhou, Mike~Zheng Shou, and Chunhua Shen.
\newblock Datasetdm: Synthesizing data with perception annotations using diffusion models.
\newblock \emph{arXiv preprint arXiv:2308.06160}, 2023{\natexlab{b}}.

\bibitem[Yang et~al.(2023{\natexlab{a}})Yang, Ma, Peng, Guo, Lin, and Yu]{yang2023bevcontrol}
Kairui Yang, Enhui Ma, Jibin Peng, Qing Guo, Di~Lin, and Kaicheng Yu.
\newblock Bevcontrol: Accurately controlling street-view elements with multi-perspective consistency via bev sketch layout.
\newblock \emph{arXiv preprint arXiv:2308.01661}, 2023{\natexlab{a}}.

\bibitem[Yang et~al.(2023{\natexlab{b}})Yang, Gao, Wang, Xu, and Xu]{yang2023mma}
Yijun Yang, Ruiyuan Gao, Xiaosen Wang, Nan Xu, and Qiang Xu.
\newblock Mma-diffusion: Multimodal attack on diffusion models.
\newblock \emph{arXiv preprint arXiv:2311.17516}, 2023{\natexlab{b}}.

\bibitem[Zhang et~al.(2023{\natexlab{a}})Zhang, Rao, and Agrawala]{zhang2023adding}
Lvmin Zhang, Anyi Rao, and Maneesh Agrawala.
\newblock Adding conditional control to text-to-image diffusion models.
\newblock In \emph{ICCV}, 2023{\natexlab{a}}.

\bibitem[Zhang et~al.(2023{\natexlab{b}})Zhang, Yang, Feng, Qin, Chen, Yu, Chen, Wang, Savarese, Ermon, et~al.]{zhang2023hive}
Shu Zhang, Xinyi Yang, Yihao Feng, Can Qin, Chia-Chih Chen, Ning Yu, Zeyuan Chen, Huan Wang, Silvio Savarese, Stefano Ermon, et~al.
\newblock Hive: Harnessing human feedback for instructional visual editing.
\newblock \emph{arXiv preprint arXiv:2303.09618}, 2023{\natexlab{b}}.

\bibitem[Zhao et~al.(2023)Zhao, Bai, Rao, Zhou, and Lu]{zhao2023unipc}
Wenliang Zhao, Lujia Bai, Yongming Rao, Jie Zhou, and Jiwen Lu.
\newblock Unipc: A unified predictor-corrector framework for fast sampling of diffusion models.
\newblock \emph{arXiv preprint arXiv:2302.04867}, 2023.

\bibitem[Zheng et~al.(2023)Zheng, Gao, and Xu]{zheng2023non}
Ziyang Zheng, Ruiyuan Gao, and Qiang Xu.
\newblock Non-cross diffusion for semantic consistency.
\newblock \emph{arXiv preprint arXiv:2312.00820}, 2023.

\bibitem[Zhili et~al.(2023)Zhili, Chen, Han, Lanqing, Xu, Li, and Kwok]{zhili2023task}
LIU Zhili, Kai Chen, Jianhua Han, HONG Lanqing, Hang Xu, Zhenguo Li, and James Kwok.
\newblock Task-customized masked autoencoder via mixture of cluster-conditional experts.
\newblock In \emph{ICLR}, 2023.

\bibitem[Zhou et~al.(2019)Zhou, Zhao, Puig, Xiao, Fidler, Barriuso, and Torralba]{ADE20K}
Bolei Zhou, Hang Zhao, Xavier Puig, Tete Xiao, Sanja Fidler, Adela Barriuso, and Antonio Torralba.
\newblock Semantic understanding of scenes through the ade20k dataset.
\newblock In \emph{IJCV}, 2019.

\bibitem[Zhou \& Kr{\"a}henb{\"u}hl(2022)Zhou and Kr{\"a}henb{\"u}hl]{zhou2022cross}
Brady Zhou and Philipp Kr{\"a}henb{\"u}hl.
\newblock Cross-view transformers for real-time map-view semantic segmentation.
\newblock In \emph{CVPR}, 2022.

\end{thebibliography}
\bibliographystyle{iclr2024_conference}

\newpage
{\centering\section*{APPENDIX}}
\appendix
\section{Object Filtering}
\begin{figure}[ht]
    \centering
    \includegraphics[width=0.98\linewidth]{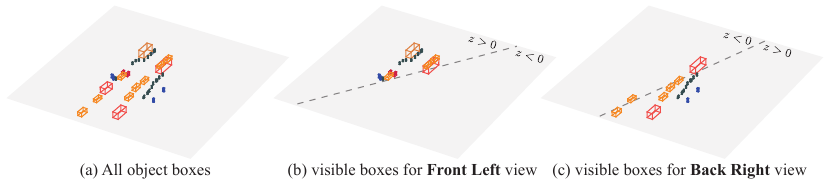}
    \caption{Bounding box filtering to each view. The dashed lines in (b-c) represent the $x$-axis of the camera's coordinates. Boxes are retained only if they have at least a point in the positive half of the $z$-axis in each camera's coordinate.}
    \label{fig:visible_box}
    \vspace{-0.2cm}
\end{figure}

In \Eqref{equ:box_filter}, 
we employ $f_{viz}$ for object filtering to facilitate bootstrap learning. We show more details of $f_{viz}$ here. Refer to \Figref{fig:visible_box} for illustration.
For the sake of simplicity, each camera's Field Of View (FOV) is not considered. Objects are defined as \textit{visible} if any corner of their bounding boxes is located in front of the camera (\ie, $z^{v_{i}}>0$) within each camera’s coordinate system. The application of $f_{viz}$ significantly lightens the workload of the bounding box encoder, evidence for which can be found in 
\Secref{sec:ablation}.

\section{More Experimental Details}\label{app:exp_setup}
\textbf{Semantic Classes for Generation.}
To support most perception models on nuScenes, we try to include semantics commonly used in most settings~\citep{huang2021bevdet,zhou2022cross,liu2022bevfusion,ge2023metabev}.
Specifically, for objects, ten categories include car, bus, truck, trailer, motorcycle, bicycle, construction vehicle, pedestrian, barrier, and traffic cone.
For the road map, eight categories include drivable area, pedestrian crossing, walkway, stop line, car parking area, road divider, lane divider, and roadblock.

\textbf{Optimization.}
We train all newly added parameters using AdamW~\citep{loshchilov2018decoupled} optimizer and a constant learning rate at $8e^{-5}$ and batch size 24 (total 144 images for 6 views) with a linear warm-up of 3000 iterations, and set $\gamma^{s}=0.2$.

\section{Ablation on Number of Attending Views}\label{app:ablation_view}
\begin{table}[ht]
    \vspace{-0.5cm}
    \captionof{table}{{Ablation on number of attending views.} Evaluation results are from CVT on the synthetic nuScenes validation set, without $M=\{0\}$ in CFG scale $=2$.}
    \label{tab:attend_ablation}
    \centering
    \begin{tabular}[b]{@{}l|ccc@{}}
        \toprule
        Attending Num. & FID $\downarrow$
        & Road mIoU $\uparrow$
        & Vehicle mIoU $\uparrow$ \\

        \midrule
        1 & \textbf{13.06} & 58.63 & 26.41\\
        2 (ours) & 14.46 & \textbf{59.31} & \textbf{27.13} \\
        5 (all) & 14.76 & 58.35 & 25.41 \\
            
        \bottomrule
    \end{tabular}
    \vspace{-0.1cm}
\end{table}

\begin{figure}[t]
    \vspace{-0.3cm}
    \centering
    \includegraphics[width=\linewidth]{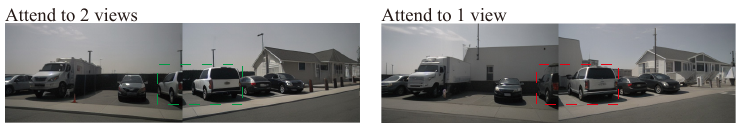}
    \vspace{-0.2cm}
    \caption{{Comparison between different numbers of attending views.} Only attending to one view results in worse multi-camera consistency.}
    \label{fig:attending-views}
    \vspace{-0.2cm}
\end{figure}
In \Tabref{tab:attend_ablation}, we demonstrate the impact of varying the number of attended views on evaluation results.
Attending to a single view yields superior FID results; the reduced influx of information from neighboring views simplifies optimization for that view.
However, this approach compromises mIoU, also reflecting less consistent generation, as depicted in \Figref{fig:attending-views}.
Conversely, incorporating all views deteriorates performance across all metrics, potentially due to excessive information causing interference in cross-attention.
Since each view has an intersection with both left and right views, attending to one view cannot guarantee consistency, especially for foreground objects, while attending to more views requires more computation.
Thus, we opt for 2 attended views in our main paper, striking a balance between consistency and computational efficiency.

\section{Qualitative Comparison with BEVGen}\label{sec:comp-bevgen}
\begin{figure}[h]
    \centering
    \includegraphics[width=0.98\linewidth]{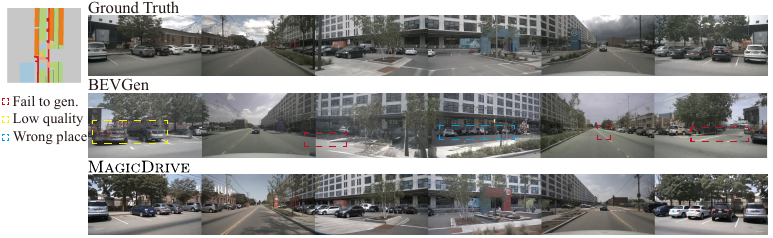}
    \caption{Qualitative comparison with BEVGen on driving scene from nuScenes validation set. We highlight some areas with rectangles to ease comparison. Compared with BEVGen, image quality of objects from \methodname is much better.
    }
    \label{fig:comparison-bevgen}
    \vspace{-0.5cm}
\end{figure}

\Figref{fig:comparison-bevgen} illustrates that \methodname generates images with higher quality compared to BEVGen~\citep{swerdlow2023street}, particularly excelling in objects.
Such enhancement can be attributed to \methodname's utilization of the diffusion model and the adoption of a customized condition injection strategy. 

\section{More results with control from different conditions}\label{sec:scene-level}
\begin{figure}[h]
    \centering
    \includegraphics[width=0.98\linewidth]{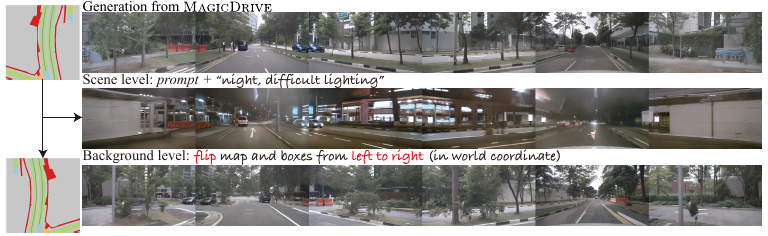}
    \caption{Showcase for scene-level control with \methodname. The scene is from the nuScenes validation set.}
    \label{fig:scene-level}
\end{figure}

\Figref{fig:scene-level} shows \textit{scene level} control (time of day) and \textit{background level} control (BEV map alterations). \methodname can effectively reflect these changes in control conditions through the generated camera views.

\newpage
\section{More Experiments with 3D Object Detection}
\begin{table}[ht]
\vspace{-0.5cm}
\centering
\caption{Comparison about support for 3D object detection model (\ie, BEVFusion). \methodname generates 272×736 images for augmentation. Results are from tests on the nuScenes validation set.
}
\label{tab:bevfusion_train_full}
\begin{tabular}{@{}cl|c|c|c|c@{}}
\toprule
\multirow{2}{*}{Epoch} & \multirow{2}{*}{Data} & \multicolumn{2}{c|}{CAM-Only} & \multicolumn{2}{c}{CAM+LiDAR} \\
\cline{3-6}
           && mAP $\uparrow$ & NDS $\uparrow$ & mAP $\uparrow$ & NDS $\uparrow$ \\
\midrule
0.5$\times$ & w/o synthetic data &
    30.21 & 32.76
    & \multirow{2}{*}{too few epochs} & \multirow{2}{*}{too few epochs} \\
    & w/ \methodname &
    33.29 {\textcolor{green}{(+3.08)}} &
    36.69 {\textcolor{green}{(+3.93)}} & & \\
   
\midrule
1$\times$ & w/o synthetic data &
    32.88 & 37.81 & 65.40 & 69.59 \\
   & w/ \methodname &
   35.40 {\textcolor{green}{(+2.52)}} &
   39.76 {\textcolor{green}{(+1.95)}} &
   67.86 {\textcolor{green}{(+2.46)}} &
   70.72 {\textcolor{green}{(+1.13)}} \\
\midrule
2$\times$ & w/o synthetic data &
    35.49 & 40.66 & 68.33 & 71.31\\
   & w/ \methodname &
   35.74 {\textcolor{green}{(+0.25)}} &
   41.40 {\textcolor{green}{(+0.74)}} &
   68.58 {\textcolor{green}{(+0.25)}} &
   71.34 {\textcolor{green}{(+0.03)}}\\
\bottomrule
\end{tabular}
\vspace{-0.2cm}
\end{table}
In \Tabref{tab:bevfusion_train_full}, we show additional experimental results on training 3D object detection models using synthetic data produced by \methodname. Given that BEVFusion utilizes a lightweight backbone (\ie, Swin-T~\citep{liu2021Swin}), model performance appears to plateau with training through 1-2$\times$ epochs (2$\times$: 20 for CAM-Only and 6 for CAM+LiDAR). Reducing epochs can mitigate this saturation, allowing more varied data to enhance the model's perceptual capacity in both settings. This improvement is evident even when epochs for 3D object detection are further reduced to 0.5$\times$. Our \methodname accurately augments street-view images with the annotations.
Future works may focus on annotation sampling and construction strategies for synthetic data augmentation.

\section{More Discussion}
\paragraph{More future work.}
Note that \methodname-generated street views can currently only perform as augmented samples to train with real data, and it is exciting to train detectors solely with generated data, which will be explored in the future.
More flexible usage of the generated street views beyond data augmentation, especially incorporation with generative pre-training~\citep{chen2023mixed,zhili2023task}, contrastive learning~\citep{chen2021multisiam,liu2022task} and the large language models (LLMs)~\citep{chen2023gaining,gou2023mixture}, is an appealing future research direction.
It is also interesting to utilize the geometric controls in different circumstances beyond 3D scenarios (\eg, multi-object tracking~\citep{li2023trackdiffusion} and concept removal~\citep{liu2023geomerasing}).

\section{Detailed Analysis on 3D Object Detection with Synthetic Data}

\begin{table}[h]
\setlength{\tabcolsep}{2.2pt}
\centering
\caption{Per-class performance comparison with BEVFusion for 3D object detection with $1\times$ setting. Results are tested on the nuScenes validation set.}
\label{tab:bevfusion_compare}
\begin{tabular}{l|c|cccccccccc}
\toprule
Data         & mAP   & car   & cone  & barrier & bus   & ped. & motor. & truck & bicycle & trailer & constr. \\
\midrule
BEVFusion & 32.88 & 50.67 & 50.46 & 48.62   & 37.73 & 35.74      & 30.40      & 27.54 & 24.85   & 15.56   & 7.28         \\
$+$ \methodname  & 35.40 & 51.86 & 53.56 & 51.15   & 40.43 & 38.10      & 33.11      & 29.35 & 27.85   & 18.74   & 9.83         \\
\midrule
Difference      &
\textcolor{green}{+2.52}  &
\textcolor{green}{+1.20}  &
\textcolor{green}{+3.10}  &
\textcolor{green}{+2.53}  &
\textcolor{green}{+2.70}  &
\textcolor{green}{+2.36}  &
\textcolor{green}{+2.71}  &
\textcolor{green}{+1.81}  &
\textcolor{green}{+3.00}  &
\textcolor{green}{+3.19}  &
\textcolor{green}{+2.55} \\
\bottomrule
\end{tabular}
\end{table}

We provide per-class AP for 3D object detection from the nuScenes validation set using BEVFusion in \Tabref{tab:bevfusion_compare}. From the results, we observe that, firstly, the improvements for large objects are significant, for example, buses, trailers, and construction vehicles. Secondly, objects with less diverse appearances, such as traffic cones and barriers, show more improvement, especially compared to trucks.
Thirdly, we note that the improvement is marginal for cars, while significant for pedestrians, motorcycles, and bicycles. This may be because the baseline already performs well for cars.
For pedestrians, motorcycles, and bicycles, even though distant objects from the ego car are generated less faithfully, \methodname can synthesize high-quality objects near the ego car, as shown in \Figref{fig:more2}-\ref{fig:more3}.
Therefore, more accurate detection of objects near the ego car contributes to improvements for these classes.
Overall, mAP improvement comes with promotion in all classes' AP, indicating \methodname can indeed help the training of perception models.

\section{More results for BEV segmentation}
BEVFusion is also capable of BEV segmentation and considers most of the classes we used in the BEV map condition. Due to the lack of baselines, we present the results in \Tabref{tab:bevfusion_seg} to facilitate comparison for future works.
As can be seen, the 272$\times$736 resolution does not outperform the 224$\times$400 resolution.
This is consistent with the results from CVT in \Tabref{tab:test} on the Road segment. Such results confirm that better map controls rely on maintaining the original aspect ratio for generation training (\ie, avoiding cropping on each side).
\begin{table}[h]
\vspace{-0.3cm}
\centering
\caption{Generation fidelity to BEV map conditions. Results are tested with BEVFusion for BEV segmentation on the nuScenes validation set.}
\label{tab:bevfusion_seg}
\begin{tabular}{@{}l|c|cc@{}}
\toprule
\multirow{2.5}{*}{Methods} & \multirow{2.5}{*}{resolution} & \multicolumn{2}{c}{mIOU for 6 classes} \\\cmidrule(l){3-4} 
            &                & CAM-only & CAM+LiDAR \\
\midrule
Oracle      & -              & 57.09    & 62.94     \\
Oracle      & 224$\times$400 & 52.72    & 58.49     \\
\midrule
\methodname              & 224$\times$400              & \textbf{30.24}     & \textbf{48.21}    \\
\methodname & 272$\times$736 & 28.71    & 47.12     \\ \bottomrule
\end{tabular}
\vspace{-0.3cm}
\end{table}

\section{Generalization of Camera Parameters}
\begin{figure}[h]
    \vspace{-0.4cm}
    \centering
    \includegraphics[width=\linewidth]{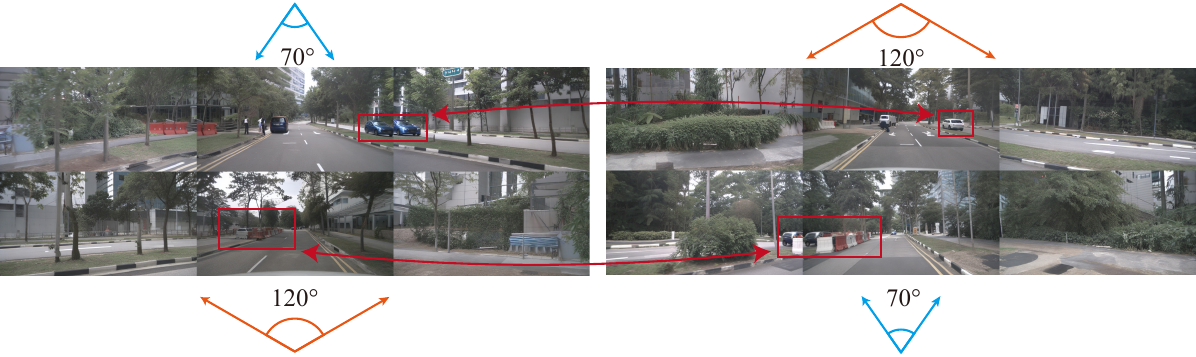}
    \caption{
    To show the generalization ability of learned camera encoding, we exchange the camera parameters between 3 front cameras and 3 back cameras. The 3D position is the same for two generations. We highlight some areas (with box boxes)}
    \label{fig:cam-generalize}
\end{figure}

To improve generalization ability, \methodname encodes raw camera intrinsic and extrinsic parameters for different perspectives. 
However, the generalization ability is somewhat limited due to nuScenes fixing camera poses for different scenes.
Nevertheless, we attempt to exchange the intrinsic and extrinsic parameters between the three front cameras and three back cameras.
The comparison is shown in \Figref{fig:cam-generalize}.
Since the positions of the nuScenes cameras are not symmetrical from front to back, and the back camera has a 120$^{\circ}$ FOV compared to the 70$^{\circ}$ FOV of the other cameras, clear differences between front and back views can be observed for the same 3D coordinates.

\newpage
\section{More Generation Results}
\begin{figure}[h]
    \centering
    \includegraphics[width=0.98\linewidth]{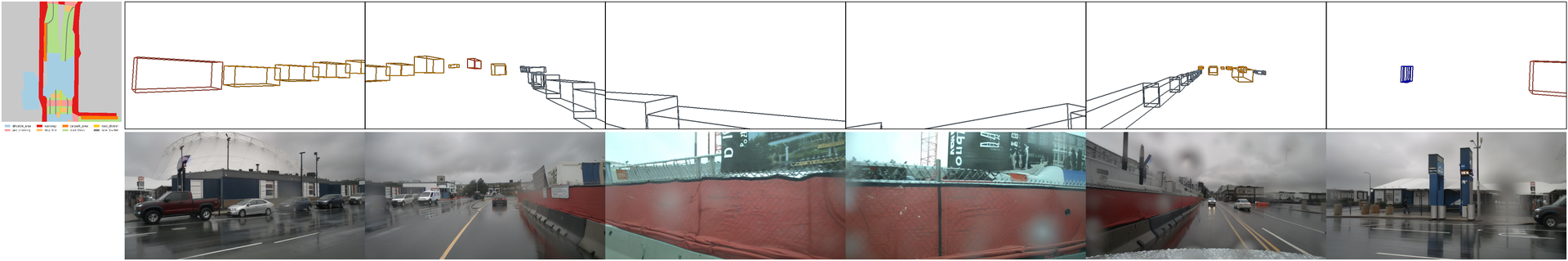}
    \includegraphics[width=0.98\linewidth]{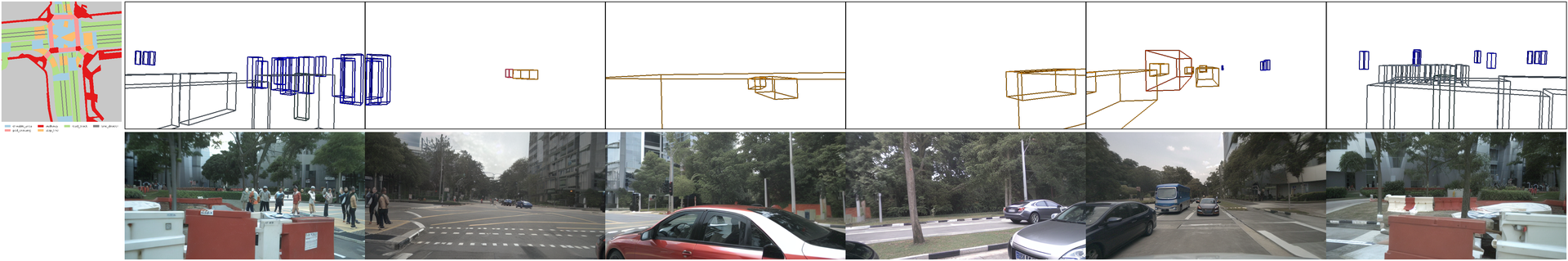}
    \includegraphics[width=0.98\linewidth]{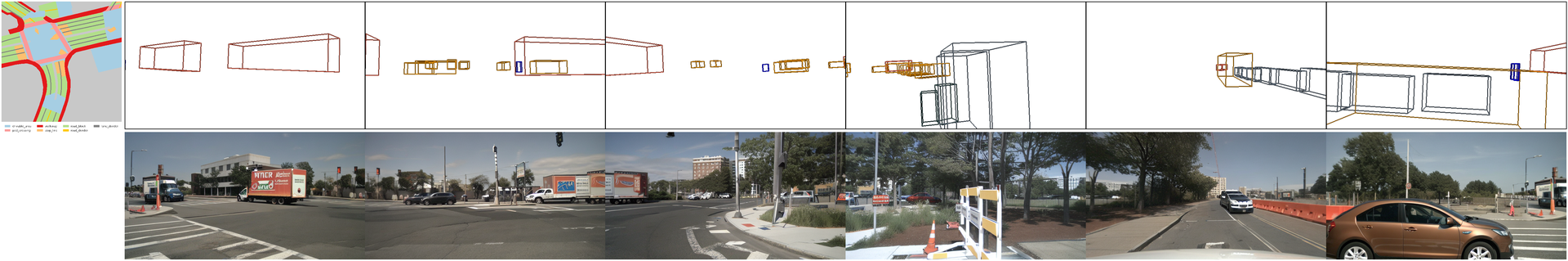}
    \caption{{Generation from \methodname with corner-case annotations}.
    }
    \label{fig:hard}
\end{figure}

\begin{figure}[ht]
    \centering
    \includegraphics[width=0.98\linewidth]{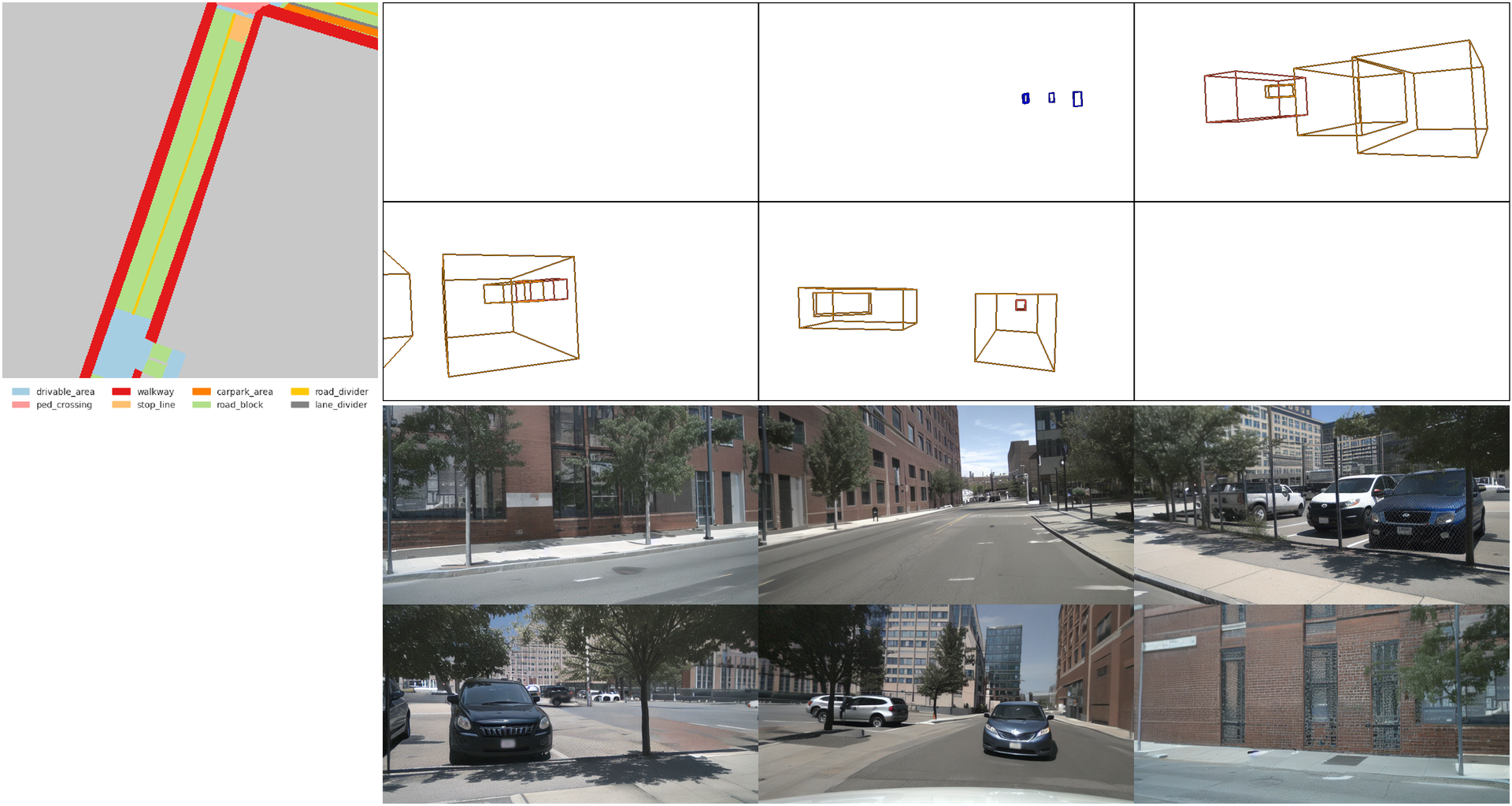}
    \includegraphics[width=0.98\linewidth]{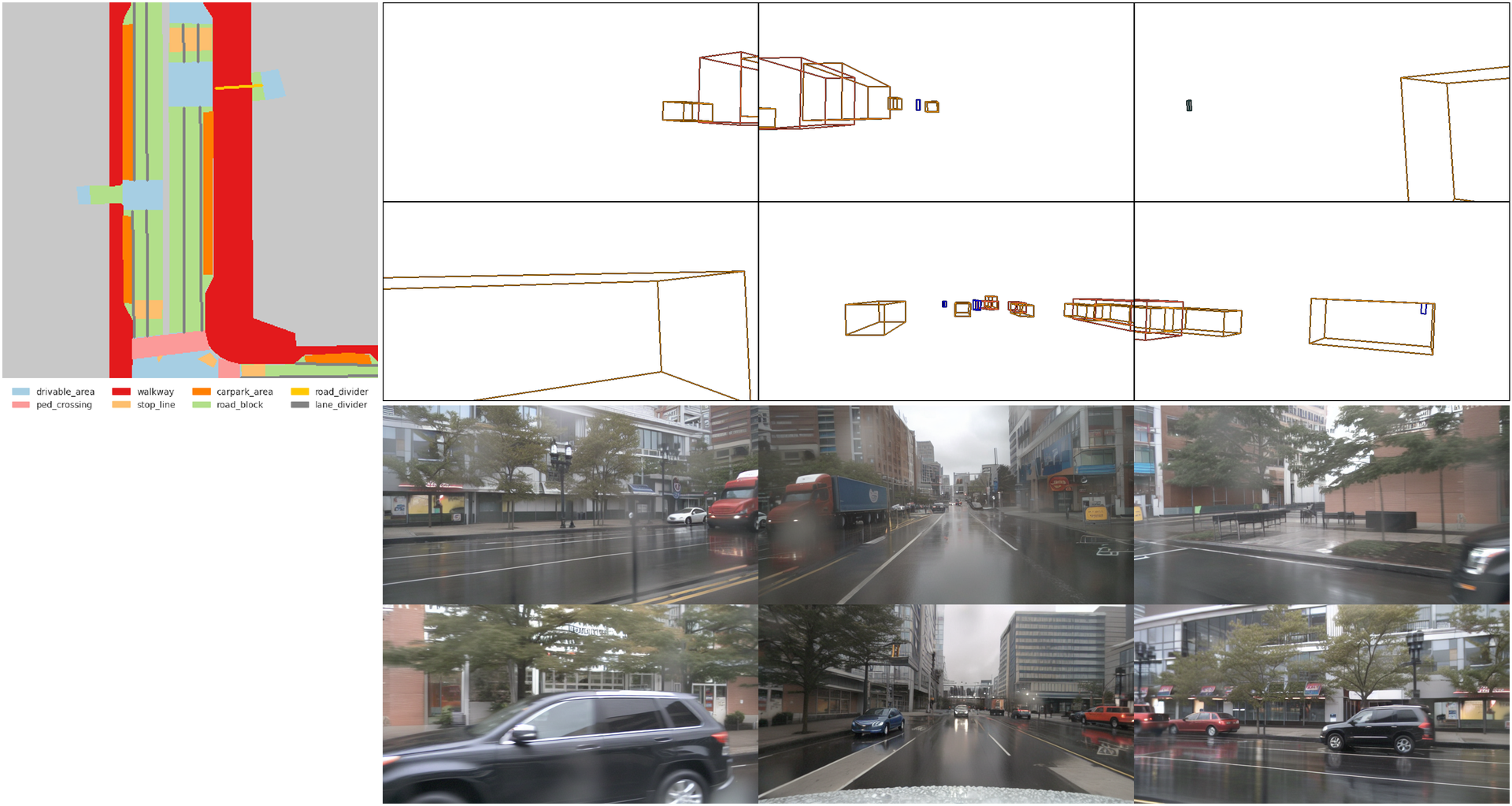}
    \includegraphics[width=0.98\linewidth]{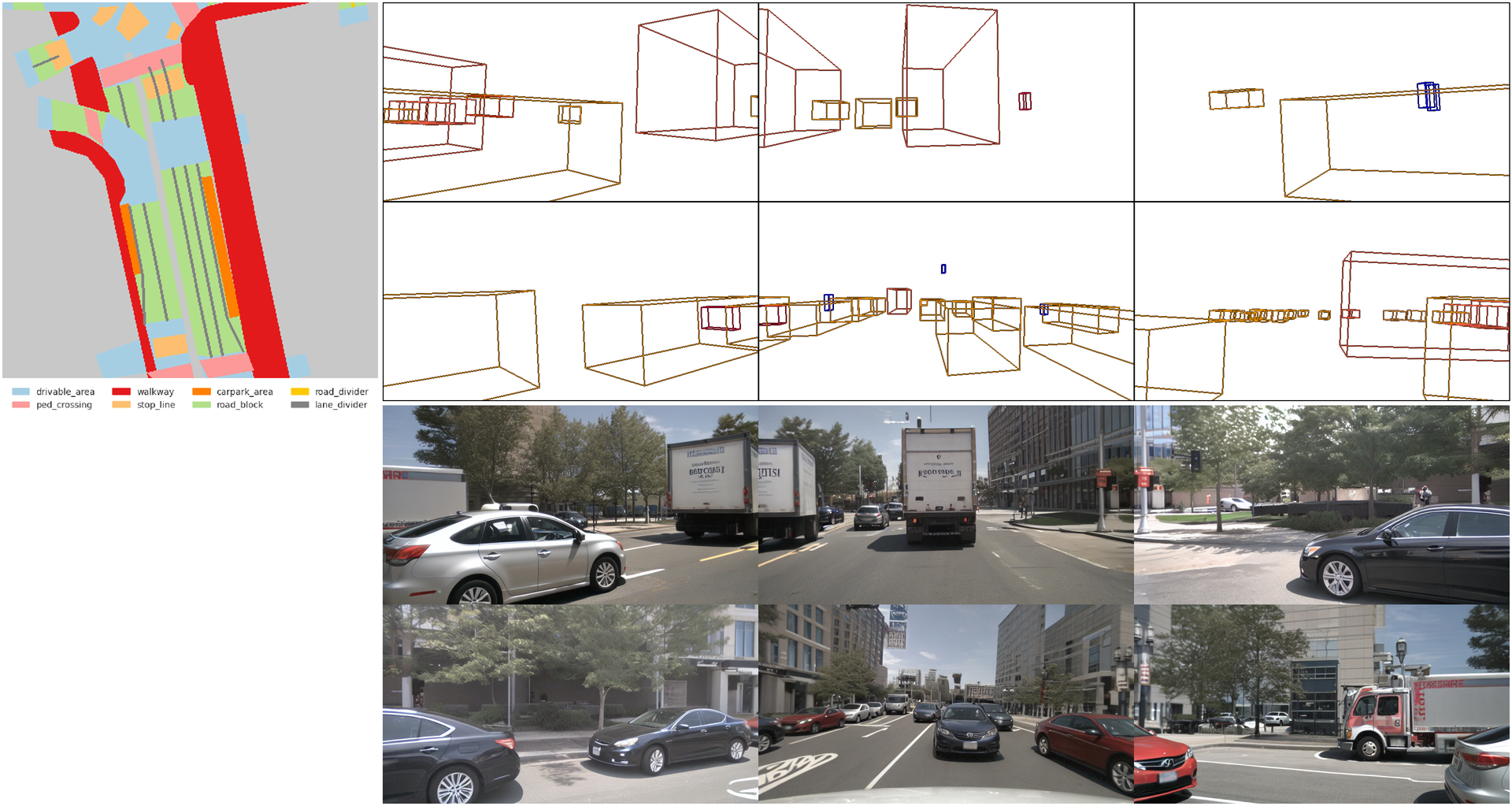}

    \caption{{Generation from \methodname with annotations from nuScenes validation set.}}
    \label{fig:more1}
\end{figure}

\begin{figure}[ht]
    \centering
    \includegraphics[width=0.98\linewidth]{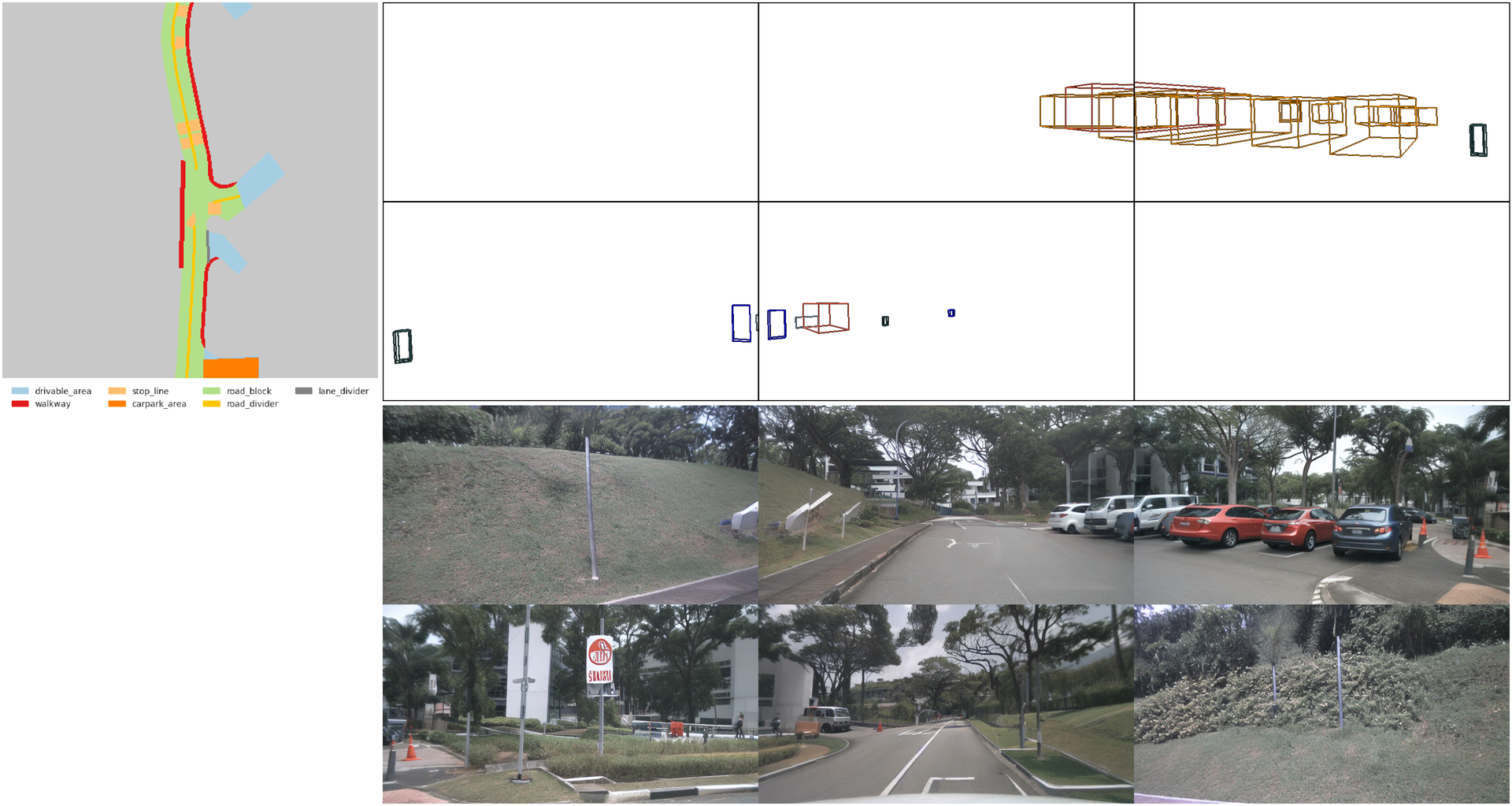}
    \includegraphics[width=0.98\linewidth]{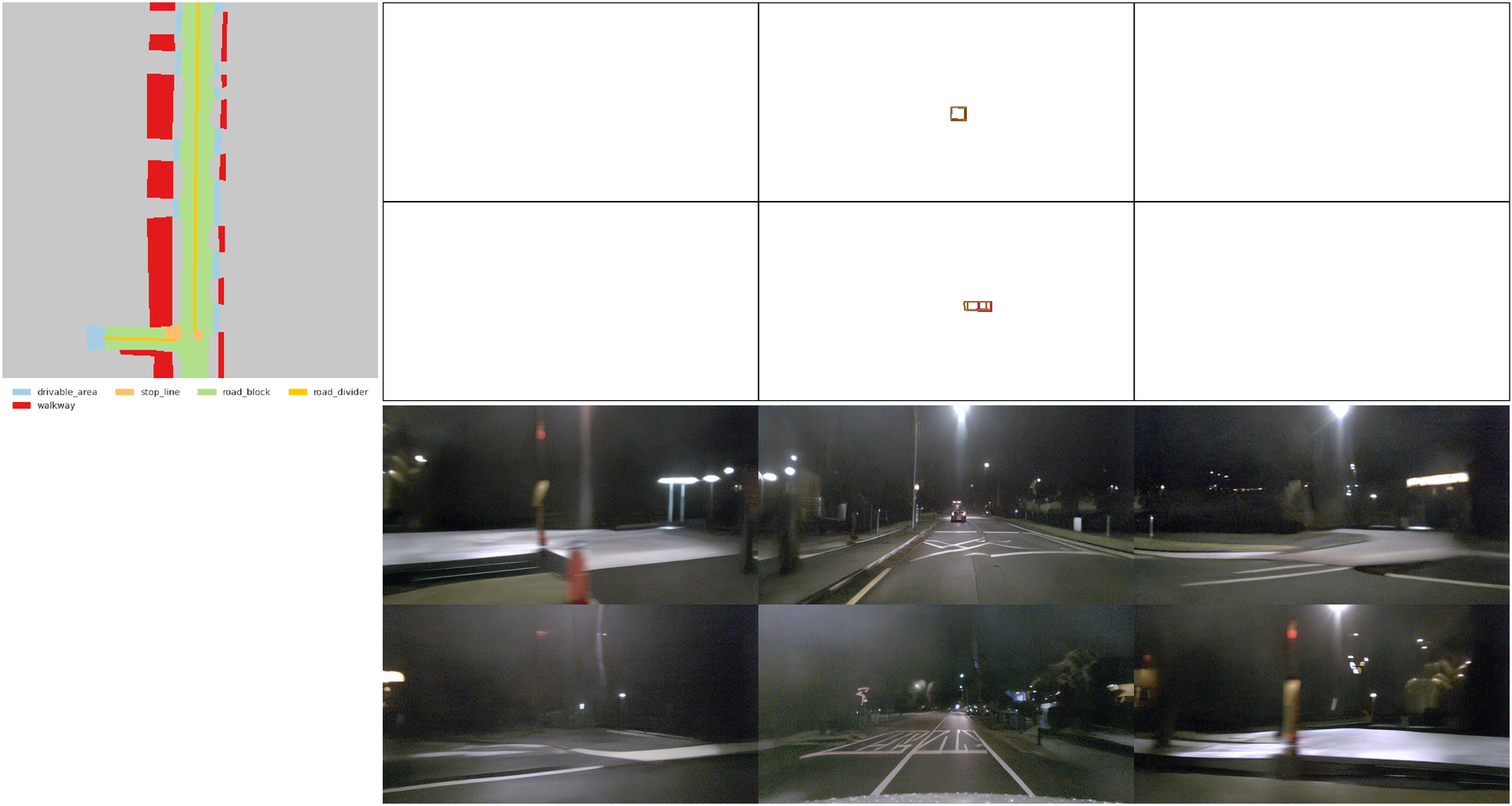}
    \includegraphics[width=0.98\linewidth]{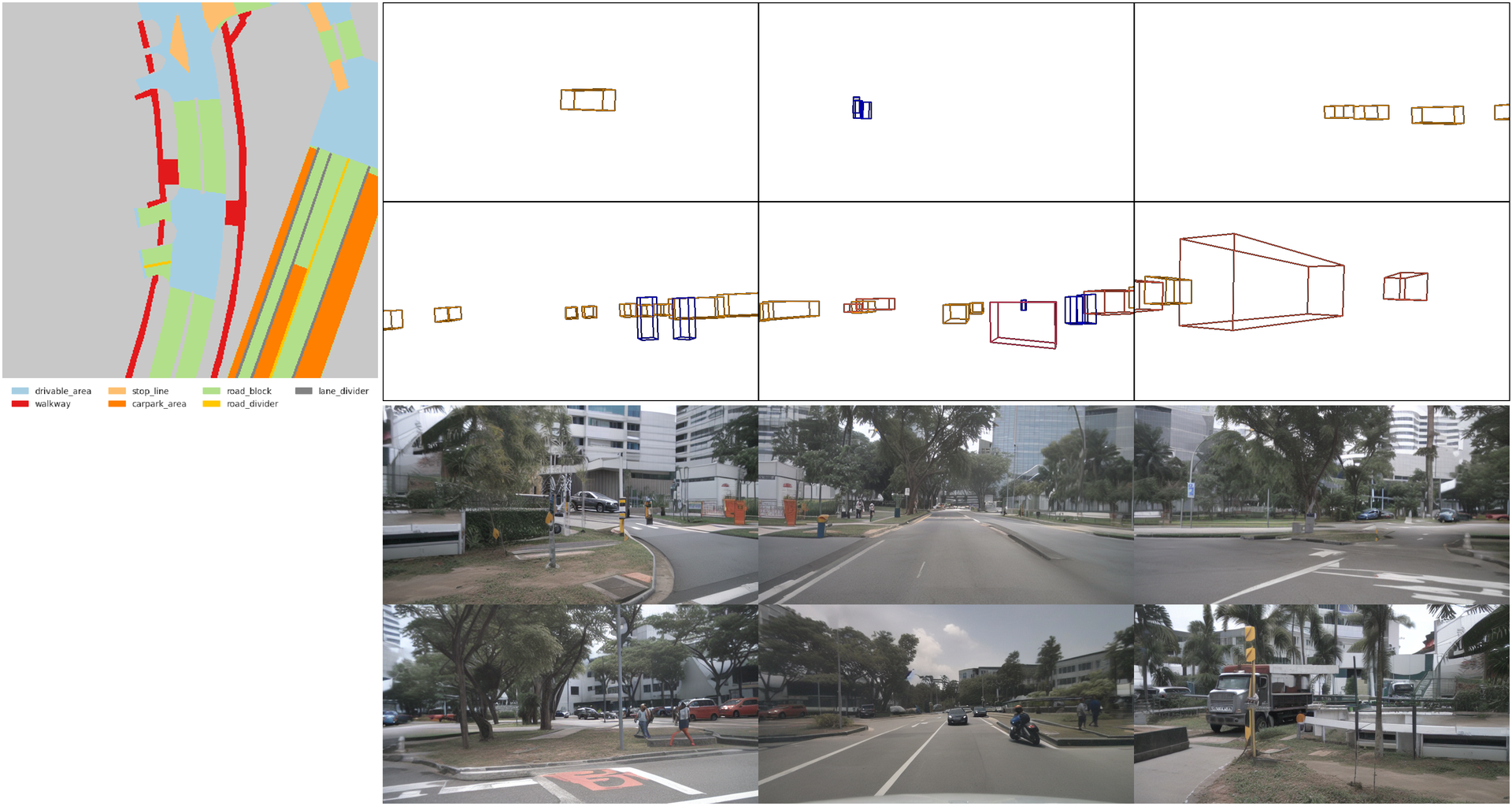}

    \caption{{Generation from \methodname with annotations from nuScenes validation set.}}
    \label{fig:more2}
\end{figure}

\begin{figure}[ht]
    \centering
    \includegraphics[width=0.98\linewidth]{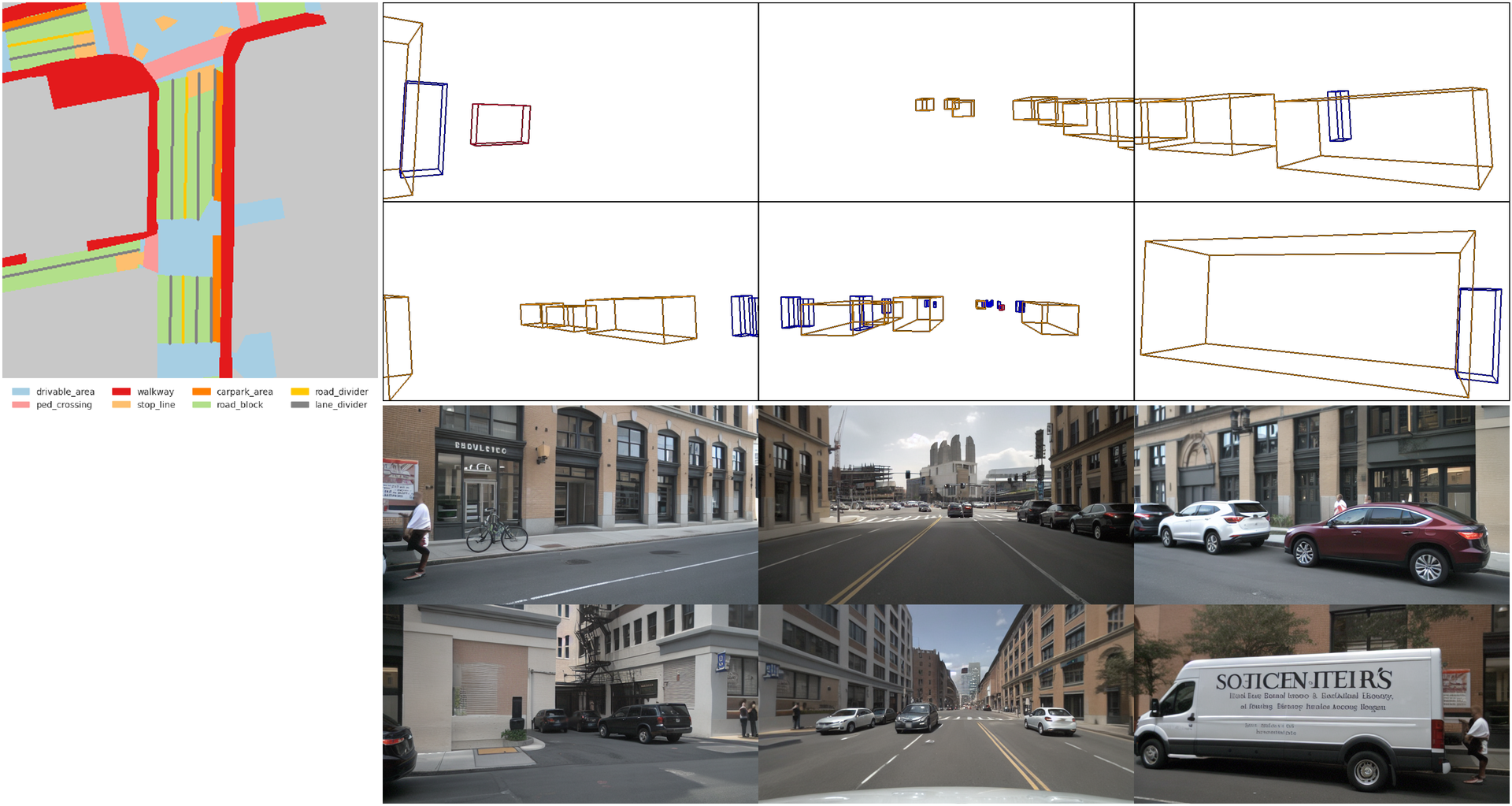}
    \includegraphics[width=0.98\linewidth]{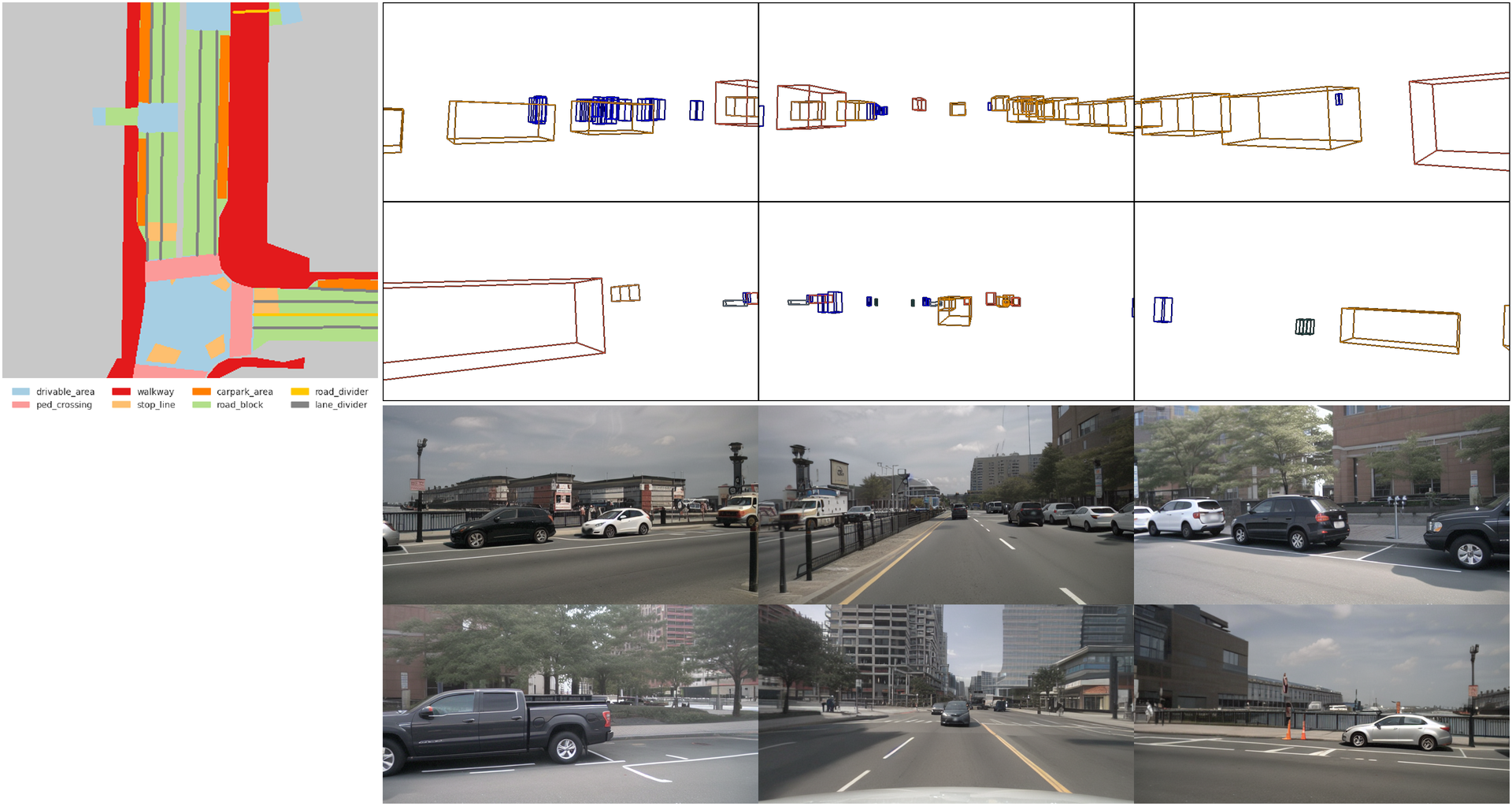}
    \includegraphics[width=0.98\linewidth]{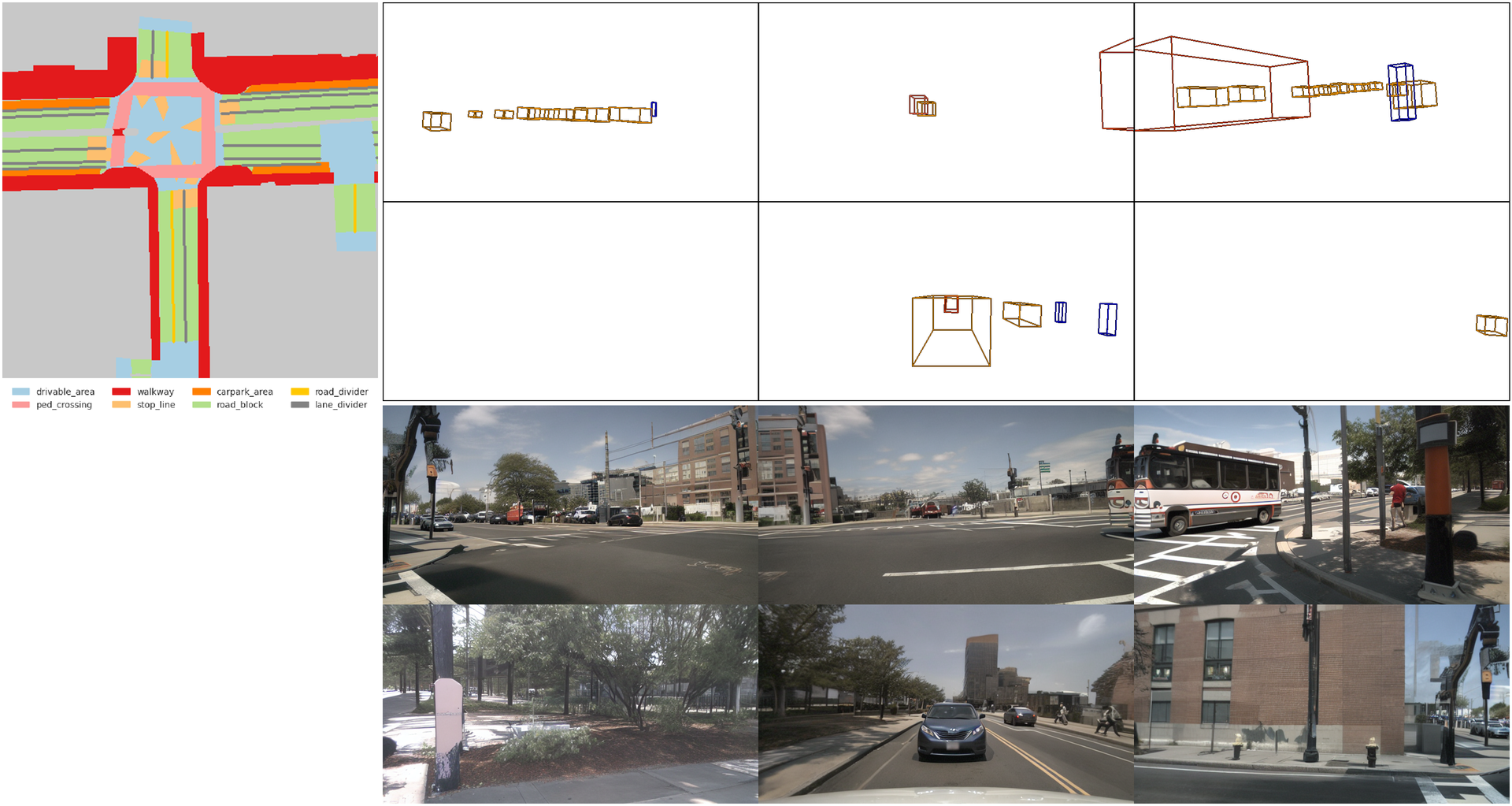}
    \caption{{Generation from \methodname with annotations from nuScenes validation set.}}
    \label{fig:more3}
\end{figure}

We show some corner-case~\citep{li2022coda} generations in \Figref{fig:hard}, and more generations 
in \Figref{fig:more1}-\Figref{fig:more3}.

\end{document}